\documentclass[runningheads]{llncs}
\usepackage{graphicx}
\usepackage{tikz}
\usepackage{comment}
\usepackage{amsmath,amssymb} 
\usepackage{color}
\usepackage{xcolor}
\usepackage{ragged2e}
\usepackage{epsfig}
\usepackage{array}
\usepackage{algorithm}
\usepackage{algpseudocode}
\usepackage{subfigure}
\usepackage{multirow}
\usepackage{hyperref}

\begin{document}
\pagestyle{headings}
\mainmatter
\def\ECCVSubNumber{100}  

\title{Segmenting Transparent Objects in the Wild} 

\titlerunning{Segmenting Transparent Objects in the Wild}

\author{Enze Xie \and
    Wenjia Wang\inst{2} \and
    Wenhai Wang\inst{3} \and
    Mingyu Ding\inst{1} \and \\
    Chunhua Shen\inst{4} \and
    Ping Luo\inst{1}
\\[.21cm]
    $^{1}$ The University of Hong Kong \quad
	$^{2}$ SenseTime Research \\
	$^{3}$ Nanjing University \quad
	$^{4}$ The University of Adelaide \\
    \tt\small \{xieenze, mingyuding\}@hku.hk,
    wangwenjia@sensetime.com, wangwenhai362@smail.nju.edu.cn, chunhua.shen@adelaide.edu.au, pluo@cs.hku.hk}

\authorrunning{E. Xie et al.}
\institute{}
\maketitle

\begin{abstract}
Transparent objects such as windows and bottles made by glass  widely exist in the real world. Segmenting transparent objects is challenging because these objects have diverse appearance inherited from the image background, making them had similar appearance with their surroundings.
Besides the technical difficulty of this task,
only a few previous datasets were specially designed and collected to explore this task and most of the existing datasets have major drawbacks. They either possess limited sample size such as merely a thousand of images without manual annotations, or they generate all images  by using computer graphics method (\emph{i.e.} not real image).
To address this important problem, this work proposes a large-scale dataset for transparent object segmentation, named Trans10K, consisting of 10,428  images of real scenarios with carefully manual annotations, which are 10 times larger than the existing datasets.
The transparent objects in Trans10K are extremely challenging due to high diversity in scale, viewpoint and occlusion.
To evaluate the effectiveness of Trans10K,
we propose a novel boundary-aware segmentation method, termed Tr\-ans\-Lab, which exploits boundary as the clue to improve segmentation of transparent objects.
Extensive experiments and ablation studies demonstrate the effectiveness of Trans10K and validate the practicality of learning object boundary in TransLab. For example, TransLab significantly outperforms 20 recent object segmentation methods based on deep learning, showing that this task is largely unsolved. We believe that both Trans10K and TransLab have important contributions to both the academia and industry, facilitating future researches and applications.
The codes and models will be released at:
\href{https://github.com/xieenze/Segment_Transparent_Objects}{\color{blue} \tt github.com/xieenze/Segment\_Transparent\_Objects}

\keywords{Transparent Objects, Dataset, Benchmark, Image Segmentation, Object Boundary}
\end{abstract}

\begin{figure}[t]
    \centering
    \includegraphics[width=1\textwidth]{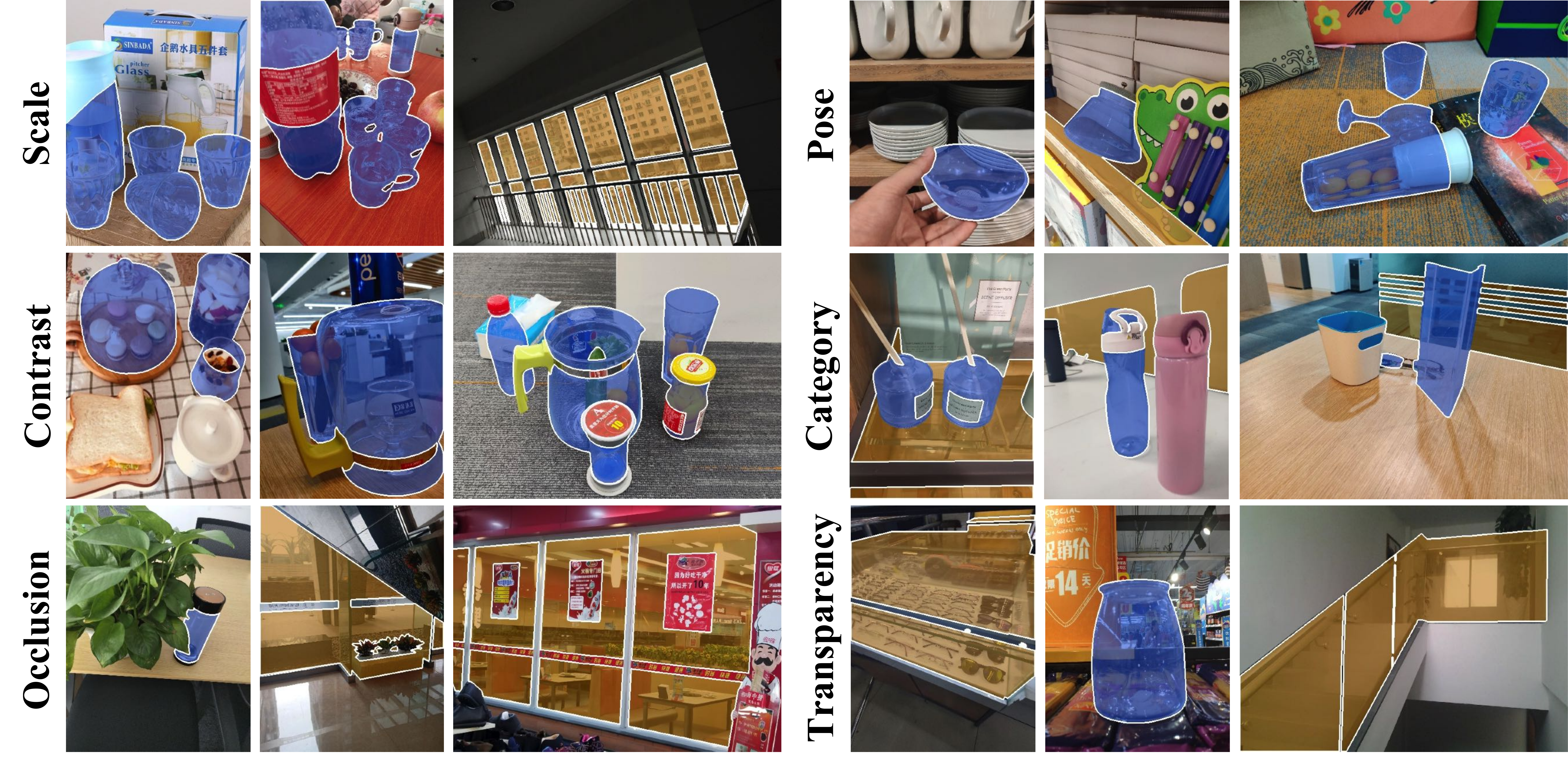}
    \caption{This work proposes the Trans10K dataset for transparent object segmentation, which has 10,428 manually-labeled images with high degree of variability in terms of scale, pose, contrast, category, occlusion and transparency. Objects are divided into two categories, thing and stuff, where \textbf{things} are small and movable objects (\emph{e.g.} bottle), while \textbf{stuff} are large and fixed (\emph{e.g.} vitrine). Example images and annotations are shown, where things and stuff are in \textbf{blue} and \textbf{brown} respectively.
    }
    \label{fig:vary}
    \end{figure}

	\section{Introduction}
	Transparent objects widely exist in the real world, such as bottles, vitrines, windows, walls and many others made by glass. Transparent objects have diverse appearance inherited from their surrounding image background, making segmenting these objects challenging.
	The task of transparent object segmentation is important because
    it has many applications.
    For example, when a smart robot operates tasks in living rooms or offices, it needs to avoid fragile objects such as glasses, vases, bowls, bottles, and jars.
    In addition, when a robot navigates in factory, supermarket and hotel, its visual navigation system needs to recognize the glass walls and windows to avoid collision.
     Although transparent object segmentation is important in computer vision,
     only a few previous datasets \cite{transcut,tomnet} were specially collected to explore this task and they have major drawbacks.
     For example, TransCut~\cite{transcut} possesses limited sample size with merely 49 images. Although TOM-Net~\cite{tomnet} has large data size of 178K images, all the images are generated by using computer graphics method by simply overlaying a transparent object on different background images, that is, the images are not real and out of the distribution of natural images.
     Meanwhile, TOM-Net provides 876 real images for test, but these images do not have manual annotations and evaluation performed by user study.

     To address the above issues, this paper proposes a novel large-scale dataset for transparent object segmentation, named Trans10K,
     containing 10,428 real-world images of transparent objects, each of which is manually labeled with segmentation mask. All images in Trans10K are selected from complex real-world scenarios that have large variations such as scale, viewpoint, contrast, occlusion, category and transparency. Trans10K has rich real images that are 10 times larger than existing datasets.

\begin{figure}[ht]
    \centering
    \scalebox{0.9}{\includegraphics[width=1\textwidth]{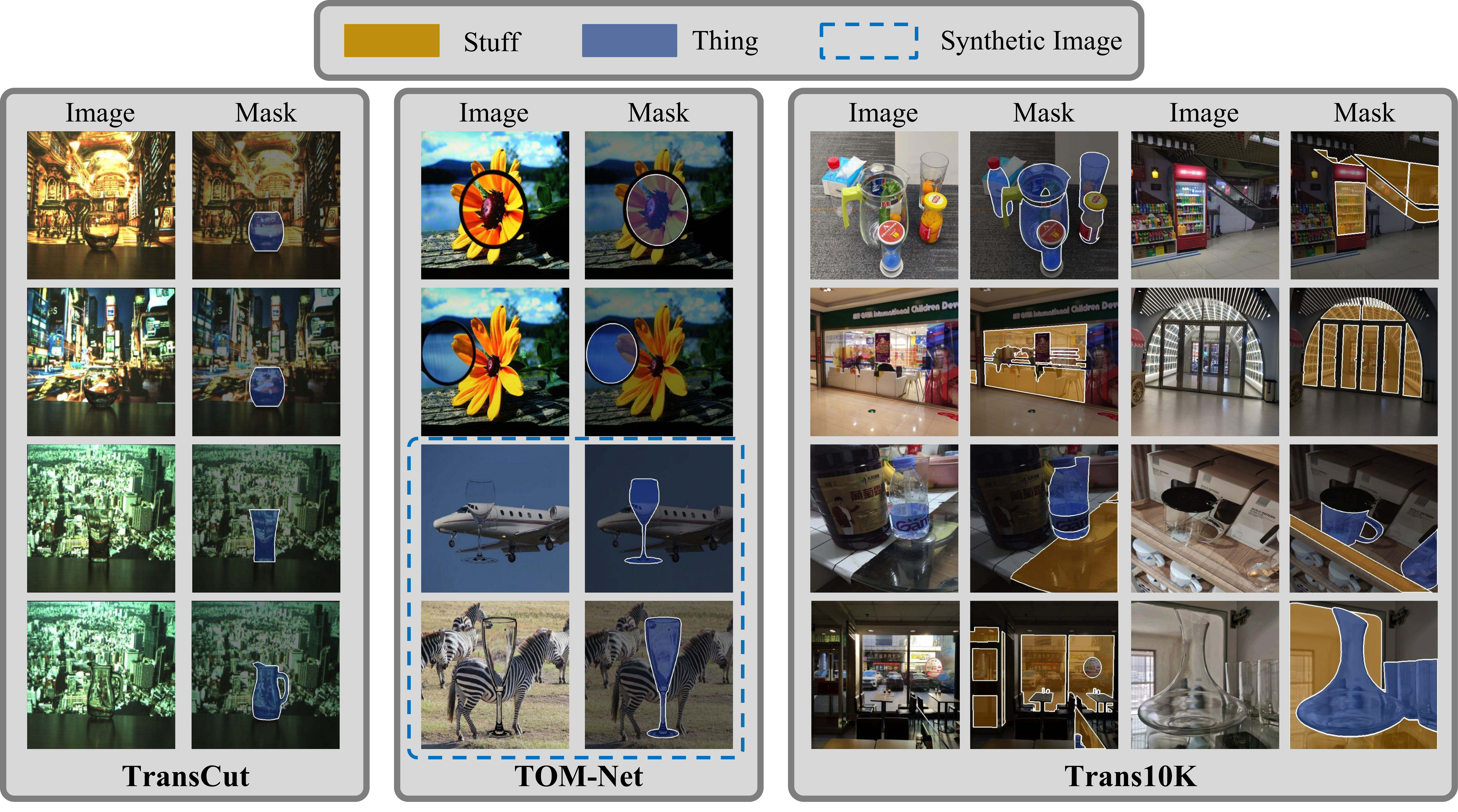}}
    \caption{Example transparent object images and masks in TransCut~\cite{transcut}, TOM-Net~\cite{tomnet}, and our Trans10K. We see Trans10K has more diverse scene and challenging viewpoint, categories, occlusion than TransCut and TOM-Net.}
    \label{fig:data_cmp}
    \end{figure}

	As shown in Table \ref{tab:data_cmp} and Fig.\ref{fig:data_cmp}, Trans10K has three main advantages compared with existing work.
	\textbf{(1)} Images in Trans10K are collected from \emph{diverse scenes} such as living room, office, supermarket, kitchen and desks, which are not covered by the existing datasets.
	\textbf{(2)} The objects in Trans10K are partitioned into {two categories}, \textbf{stuff} and \textbf{things}. The transparent stuff are fixed and large object such as wall and window.
	Segmenting stuff objects are useful for many applications, such as helping robots avoid collision during navigation.
	In contrast, the transparent things are small and moveable such as bottles.
\textbf{(3)} As shown in Table \ref{tab:data_cmp}, images in Trans10K are divided into a training set of 5000 images, a validation set of 1000 images and a test set of 4428 images. Both the validation and the test sets contain two subsets, \emph{easy and hard}. Although the overall benchmark is challenging,
    we carefully select a hard subset for validation and test to further evaluate different segmentation algorithms.
    These hard subsets may expose more flaws of semantic segmentation algorithms so as to improve the performance of transparent object segmentation.

    \begin{table}[t]
    \center
	\caption{\small{\textbf{Comparisons} between Trans10K and previous transparent object datasets, where ``Syn'' represents synthetic images using computer graphics method,  ``Thing'' represents small and movable objects, ``Stuff'' are large and fixed objects, and ``MCC'' denotes Mean Connected Components in each image.
	Trans10K is much more challenging than prior arts in terms of all characteristics presented in this table.
	}}
    \scalebox{0.7}{
    \begin{tabular}{p{60pt}<{\centering}|p{26pt}<{\centering}|p{26pt}<{\centering}|p{30pt}<{\centering}|p{26pt}<{\centering}|p{26pt}<{\centering}|p{26pt}<{\centering}|p{26pt}<{\centering}|p{26pt}<{\centering}|p{30pt}<{\centering}|p{26pt}<{\centering}|p{26pt}<{\centering}|p{40pt}<{\centering}|p{35pt}<{\centering}}
\hline
\multicolumn{1}{c|}{\multirow{3}{*}{\large Datasets}} & \multicolumn{4}{c|}{\rule{0pt}{11pt}Image Number}             &
\multicolumn{3}{c|}{Objects}               & \multicolumn{3}{c|}{Height (pixels)}                                              & \multicolumn{3}{c}{Properties}                                                                      \\ \cline{2-14}  
\multicolumn{1}{c|}{}                   & \multicolumn{2}{c|}{\rule{0pt}{11pt}Train}  & \multirow{2}{*}{Val} & \multirow{2}{*}{Test}  &  \multirow{2}{*}{\textbf{Num}}      & \multirow{2}{*}{Thing }              & \multirow{2}{*}{Stuff }                & \multirow{2}{*}{\textless{}1K }          & \multirow{2}{*}{1K-2K}     & \multirow{2}{*}{\textbf{\textgreater{}2K}} & \multirow{2}{*}{\textbf{MCC}} & \multirow{2}{*}{Occlusion} & \multirow{2}{*}{Contrast} \\\cline{2-3}
\multicolumn{1}{c|}{}       &\rule{0pt}{11pt} Real  & Syn &   &    &           &                &        &     &  &  & &  \\
\hline\hline
 {\rule{0pt}{11pt}}   TransCut~\cite{transcut}      & 0 & 0&0 & 49 & 7  &  $\surd$  &  $\times$  &   49 &  0  &    0   &             1.14     &             $\times$         &   $\times$                     \\ 
 \rule{0pt}{11pt} TOM-Net~\cite{tomnet}      &  0 &178K   &  900 & 876 &18&  $\surd$  &  $\times$   &   178K &   876  &   0    &    1.33              &                $\times$      &     $\times$                   \\ 

  {\rule{0pt}{11pt}}    \textbf{Trans10K}    &  5000  & 0  & 1000 &  4428  & 10K$+$ &  $\surd$  &  $\surd$  &  1544  &  593  &      8291 &      3.96            &             $\surd$        &    $\surd$                   \\ \hline
\end{tabular}
    }

	\label{tab:data_cmp}
    \end{table}

    \begin{figure}[t]
    \centering
    \scalebox{0.9}{\includegraphics[width=1\textwidth]{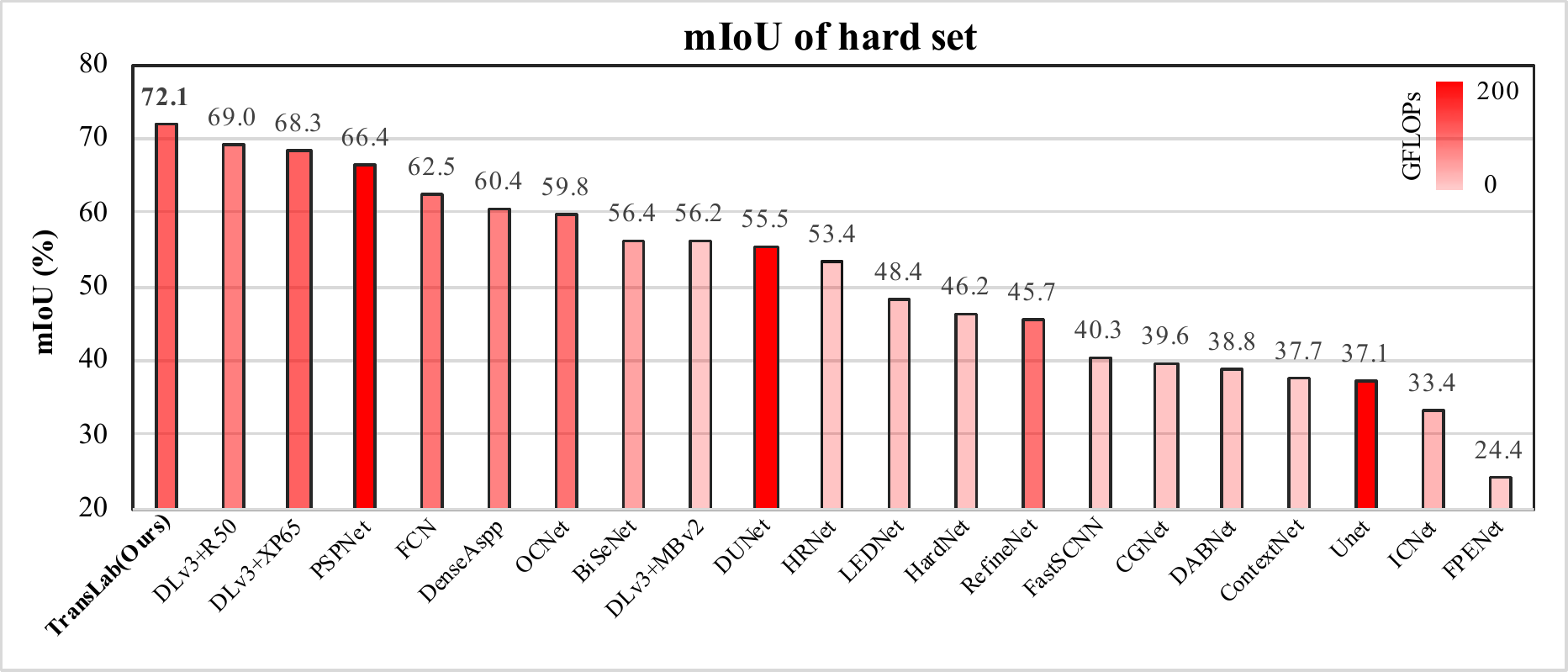}}
    \caption{\small{\textbf{Comparisons} between TransLab and 20 representative semantic segmentation methods. All methods are trained on Trans10K. mIoU on hard set of Trans10K is chosen as the metric. Deeper color bar indicates methods with larger FLOPs. `DLv3+' denotes DeepLabV3+.
    } }
    \label{fig:21bar}
    \end{figure}

	Besides Trans10K, we carefully design a boundary-aware object segmentation method, termed \textbf{TransLab}, which is able to ``\textbf{L}ook \textbf{a}t \textbf{b}oundary'' to improve transparent object segmentation.
    As shown in Fig.\ref{fig:21bar}, we train and evaluate 20 existing representative segmentation methods \cite{pspnet,icnet,dunet,deeplab,denseaspp,refinenet,fcn} on Trans10K, and found  that simply applying previous methods to this task is not sufficient.
    For instance, although DeepLabV3+~\cite{deeplabv3+} is the state-of-the-art semantic segmentation method,
    it ignores the boundary information, which might be suitable for common object segmentation but not for transparent object. As a result, its mIoU is suboptimal compared to TransLab (69.0 \emph{versus} 72.1), where
    the boundary prediction in TransLab is helpful due to the high contrast at the edges but diverse appearance inside a transparent object.

    Specifically, TransLab has two streams including a regular stream for transparent content segmentation and a boundary stream for boundary prediction.
    After these two streams, a Boundary Attention Module~(BAM) is devised to use the boundary map to attend both high-level and low-level features for transparent object segmentation.

    To summarize, the main \textbf{contributions} of this work are three-fold. \textbf{(1)} A large-scale
    transparent object segmentation dataset, Trans10K, is collected and labeled. It is 10 times larger and more challenging than previous work. \textbf{(2)} A boundary-aware approach for segmenting transparent objects,
    TransLab, is proposed to validate the effectiveness of Trans10K and the boundary attention module. TransLab surpasses 20 representative segmentation approaches by training and testing them on Trans10K for fair comparisons. For instance, TransLab outperforms DeeplabV3+, a state-of-the-art semantic segmentation method, by over 3\% mIoU. \textbf{(3)} Extensive ablation studies and benchmarking results are presented by using Trans10K and TransLab to encourage more future research efforts on this task. The data of Trans10K and trained models are released.

	\section{Related Work}
	\textbf{Semantic Segmentation.}
	Most state-of-the-art algorithms for semantic segmentation are predominantly based on CNNs. Earlier approaches~\cite{fcn,crf} transfer classification networks to fully convolutional networks (FCNs) for semantic segmentation, which is in an end-to-end training manner. Several works~\cite{deeplab,lin2016efficient,zheng2015conditional} propose to use structured prediction modules such as conditional random fields (CRFs) on network output for improving the segmentation performance, especially around object boundaries.
	To avoid costly DenseCRF, the work of~\cite{chen2016semantic} uses fast domain transform  filtering on network output while also predicting edge maps from intermediate CNN layers.
    More recently, dramatic improvements in performance and inference speed have been driven by new architectural designs. For example, PSPNet~\cite{pspnet} and DeepLab~\cite{deeplab,deeplab2} proposed a feature pyramid pooling module that incorporates multiscale context by aggregating features at multiples scales.
    Some works~\cite{gadde2016superpixel,liu2017learning,nonlocal} propose modules that use learned pixel affinities for structured information propagation across intermediate CNN representations.

	\textbf{Transparent Object Segmentation.}
    TransCut~\cite{transcut} propose an energy function based on LF-linearity and occlusion detection from the 4D light-field image is optimized to generate the segmentation result.
    TOM-Net~\cite{tomnet} formulate transparent object matting as a refractive ﬂow estimation problem. This work proposed a multi-scale encoder-decoder network to generate a coarse input, and then a residual network refines it to a detailed matte.
    Note that TOM-Net needs a refractive flow map as label during training, which is hard to obtain from the real world, so it can \textbf{only} rely on synthetic training data.

	\textbf{Transparent Object Datasets.}
    TOM-Net~\cite{tomnet} proposed a dataset containing 876 real images and 178K synthetic images which are generated by POV-Ray. Only 4 and 14 objects are repeatedly used in the synthetic and real data. Moreover, the test set of TOM-Net do not have mask annotation, so one cannot evaluate his algorithm quantitatively on it.
    TransCut~\cite{transcut} is proposed for the segmentation of transparent objects. It only contains 49 images. However, only 7 objects, mainly bottles and cups, are repeatedly used. The images are capture by 5$\times$5 camera array in 7 different scenes. So the diversity is very limited.

    Most of the background of synthetic images are chosen randomly, so the background and the objects are not semantically coordinated and reasonable. The transparent objects are usually in a unreasonable scene, \emph{e.g.} a cup flying with a plane. Furthermore, the real data always lack in scale and complexity.

	\section{Trans10K Dataset and Annotation}
	To tackle the transparent object segmentation problem, we build the first large-scale dataset, named Trans10K. It contains more than 10k pairs of images with transparent objects and their corresponding manually annotated masks,
	which is over 10$\times$ larger than existing real transparent object datasets.
	\subsection{Data description}

	The Trans10K dataset contains 10,428 images, with two categories of transparent objects:
	(1) Transparent things such as cups, bottles and glass, locating these things can make robots easier to grab objects.
	(2) Transparent stuff such as windows, glass walls and glass doors. It can make robots learn to avoid obstacles and avoid hitting these stuff.
    As shown in Table.~\ref{tab:trans}, 5000, 1000 and 4428 images are used for train, validation and test, respectively.
    Specifically, we keep the same ratio that these two fine-grained categories in train, validation and test set.
    The images are manually harvested from the internet, image library like google OpenImage~\cite{openimage} and our own data captured by phone cameras.
    As a result, the distribution of the images is various, containing different scales, born-digital, perspective distortion glass, crowded and so on.
    In summary, to our best knowledge, Trans10K is the largest real dataset focus on transparent object segmentation in the wild.
    Moreover, due to fine-grained categories and high diversity, it is challenging enough for existing semantic segmentation methods.

	\subsection{Annotation}
	The transparent objects are manually labeled by ourselves with our labeling tool. The way of annotation is the same with semantic segmentation datasets such as ADE20K.
    We set the background with 0, transparent things with 1 and transparent stuff with 2.
    Here are some principles:
    (1) Only highly transparent objects are annotated, other semi-transparent objects are ignored.
    Although most transparent objects are made of glass in our dataset, we also annotate those made of other materials such as plastic if they satisfy the attribute of transparent.
    (2) If there are things in front of the transparent objects, we will not annotate the region of the things. Otherwise, if things are behind transparent objects, we will annotate the whole region of transparent objects.
    (3) We further divide the validation set and test set into two parts, easy and hard according to the difficulty.  The detail is shown in Section.~\ref{complexity}.

	\subsection{Dataset Complexity}
	\label{complexity}

\begin{table}[ht]
\center
\caption{
	Image statistics of \textbf{Trans10K}. “MCC” denotes Mean Connected Components in each image.}
\scalebox{0.8}{
    \begin{tabular}{p{45pt}<{\centering}p{40pt}<{\centering}|p{40pt}<{\centering}|p{50pt}<{\centering}|p{50pt}<{\centering}|p{70pt}<{\centering}|p{40pt}<{\centering}}
\hline
\multicolumn{2}{c|}{\multirow{2}{*}{\rule{0pt}{11pt}Dataset}}   & \multicolumn{4}{c|}{\rule{0pt}{11pt}Image Number} &\multirow{2}{*}{MCC} \\  \cline{3-6}
\rule{0pt}{11pt}& & All  & Only things & Only stuff & Containing both \\ \hline
\multicolumn{2}{c|}{\rule{0pt}{11pt}Train}        & 5000 & 2845                   & 2047                  & 108      &    3.87   \\ \hline
\multicolumn{1}{c|}{\multirow{2}{*}{Validation}} & \rule{0pt}{11pt}easy & 788  &         490          &290               &      8      &  3.31 \\ \cline{2-7} 
\multicolumn{1}{c|}{}              & \rule{0pt}{11pt}hard & 212  &            82         &        118          &   12   &    5.20    \\ \hline
\multicolumn{1}{c|}{\multirow{2}{*}{Test}} & \rule{0pt}{11pt}easy &  3491 &         2255        &      1222       &  14    &  3.15  \\ \cline{2-7} 
\multicolumn{1}{c|}{}                  & \rule{0pt}{11pt}hard & 937 &              337      &      549            &    51    &  6.29    \\ \hline
\end{tabular}
    }
	\label{tab:trans}
\end{table}

\begin{figure*}[htb]
    \centering
        \subfigure[Examples of easy cases. With regular shapes, less occlusion and contrast.         ]{\includegraphics[width=4.5in]{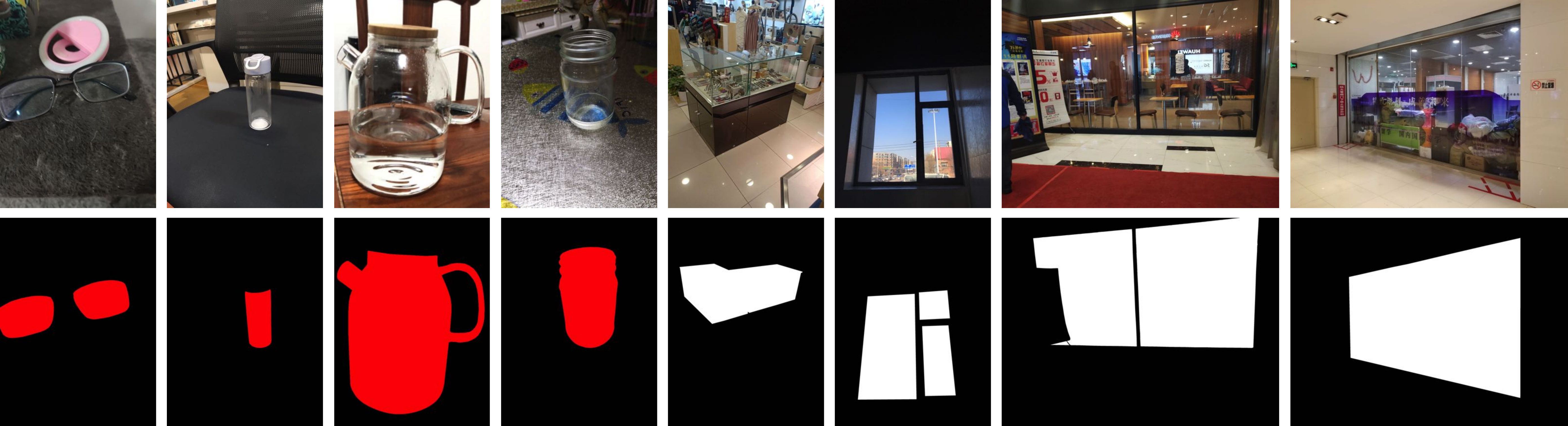}}
        \subfigure[Examples of hard cases. With irregular shapes, more occlusion and contrast.
        ]{\includegraphics[width=4.5in]{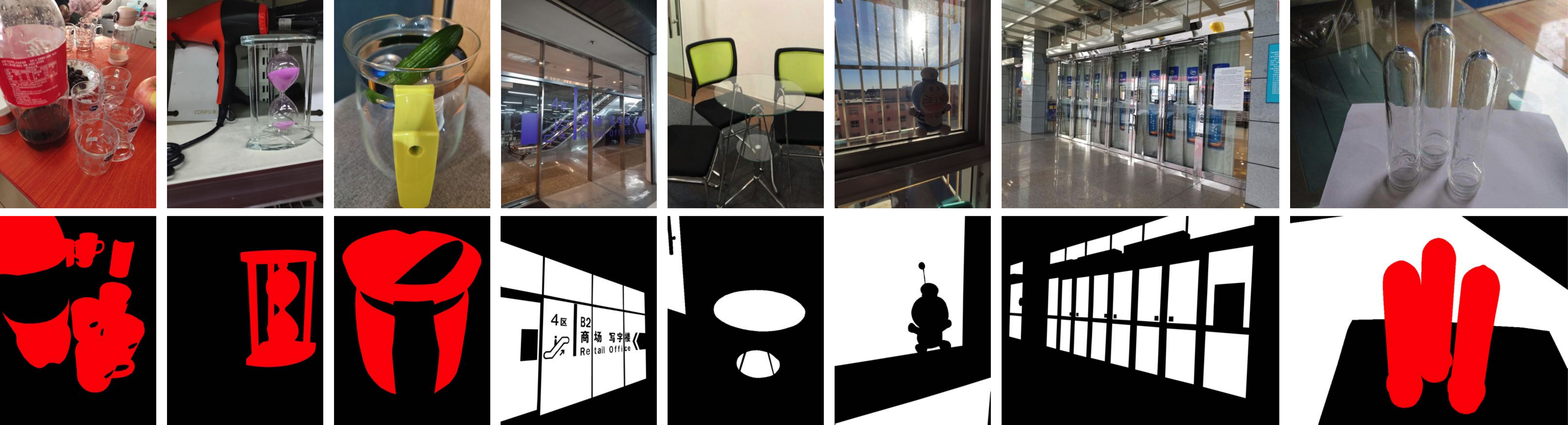}}
    \caption{Comparisons of easy and hard samples in Trans10K. \textbf{Red} represents things and \textbf{white} represents stuff. Best viewed in color.}
    \label{fig:easy_hard}
    \end{figure*}

	Our dataset is diversified in scale, category, shape, color and location.
    We find that the segmentation difficulty varies due to these factors. So we define the easy and hard attribute of each image. The statistics are shown in Table.~\ref{tab:trans}. The detailed principles are shown as below:

    \textbf{Easy}~(Fig.~\ref{fig:easy_hard} (a)):
       (1) Less numbered. \emph{e.g.} most images contain a single object of the same category.
       (2) Regular shaped. \emph{e.g.} there is nearly no occlusion like posters over the transparent objects and their shape is regular such as circle.
       (3) Salient. \emph{e.g.} the transparent objects are salient and easy to figure out due to the conspicuous reflection and refraction light.
       (4) Simply displayed. \emph{e.g.} the transparent objects are located at a center position and spatially isolated to each other clearly.

   \textbf{Hard}~(Fig.~\ref{fig:easy_hard} (b)):
       (1) More numbered. \emph{e.g.} the images contain multiple transparent objects of different categories.
       (2) Irregular shaped. \emph{e.g.} their shape is strange and without a regular pattern.
       (3) High Transparency. \emph{e.g.} they are hard to figure out because they are of very high transparency and clean, even hard for people to figure.
       (4) Complexly displayed. \emph{e.g.} they are located randomly and heavily occluded in a crowd scene. One transparent objects can cover, or contain part of another.

	Fig.~\ref{fig:easy_hard} (a) and Fig.~\ref{fig:easy_hard} (b) shows that our dataset contain abundant category distribution. To our knowledge, our dataset contains at least 20 different categories of objects, including but not limited to:
    \textbf{Stuff} such as bookshelf, showcase, freezer, window, door, glass table, vivarium, floor window, glass ceiling, glass guardrail, microwave oven and electric roaster.
    \textbf{Things} such as eyeglass, cup, bow, bottle, kettle, storage box, cosmetics, toys, glass chandelier.
    The abundant categories contain the most common transparent objects in the real world. More visualization can be found in the supplementary materials.

    Fig.~\ref{fig:statistic} displays the statistic information of the Trans10K dataset.
    (a) is the distribution of area ratio of connected components in each image, ranging from 0 to 1.0.
    (b) is the number of connected components of things and stuff in each image.
    (c) is the distribution of the image resolution of the train and validation+test set. The horizontal axis is the resolution (million pixels) of each image calculated by width $\times$ height.
    (d) is the distribution of the object location of the whole dataset. It shows that the stuff is more uniformly distributed while things tend to cluster near the center of the image. This is reasonable because the stuff tends to occupy the majority of the images.

   \begin{figure}[ht]
    \centering
    \subfigure[]{\includegraphics[width=1.1in]{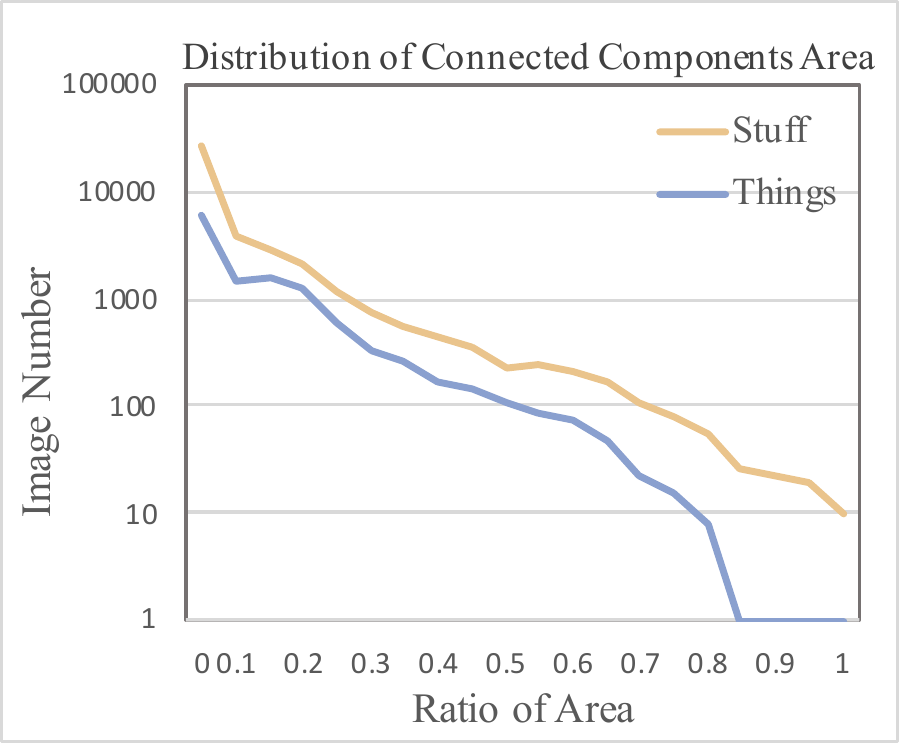}}
    \subfigure[]{\includegraphics[width=1.1in]{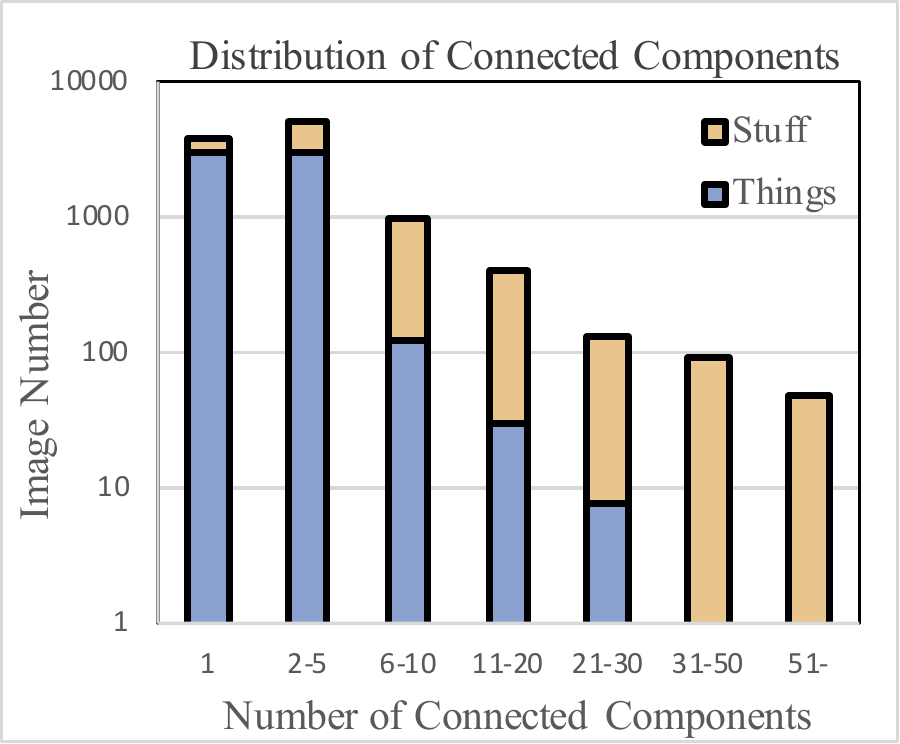}}
    \subfigure[]{\includegraphics[width=1.111in]{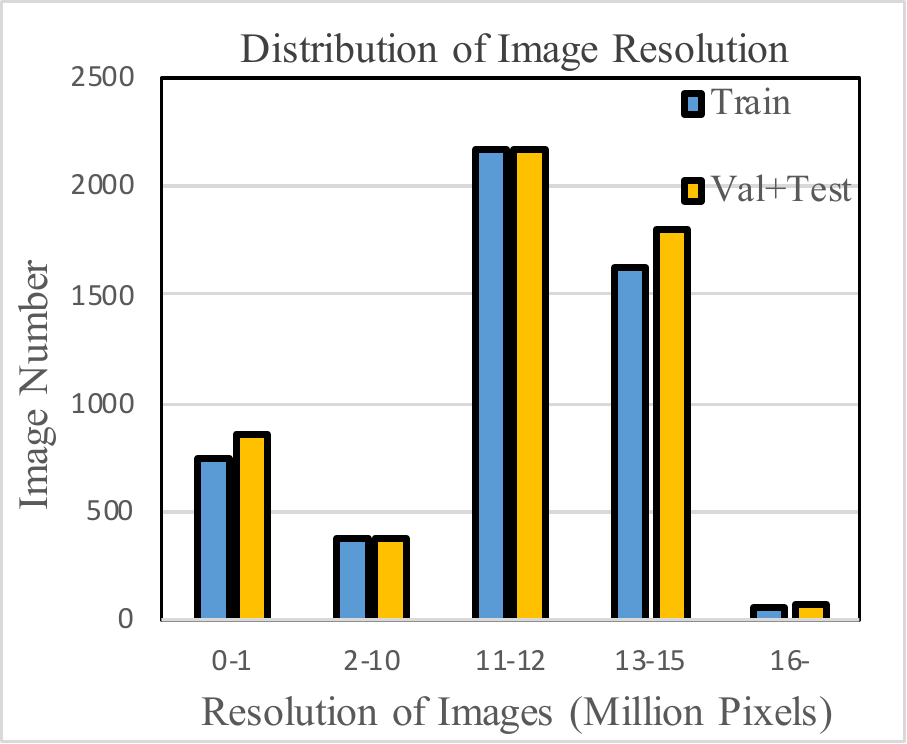}}
    \subfigure[]{\includegraphics[width=1.1in]{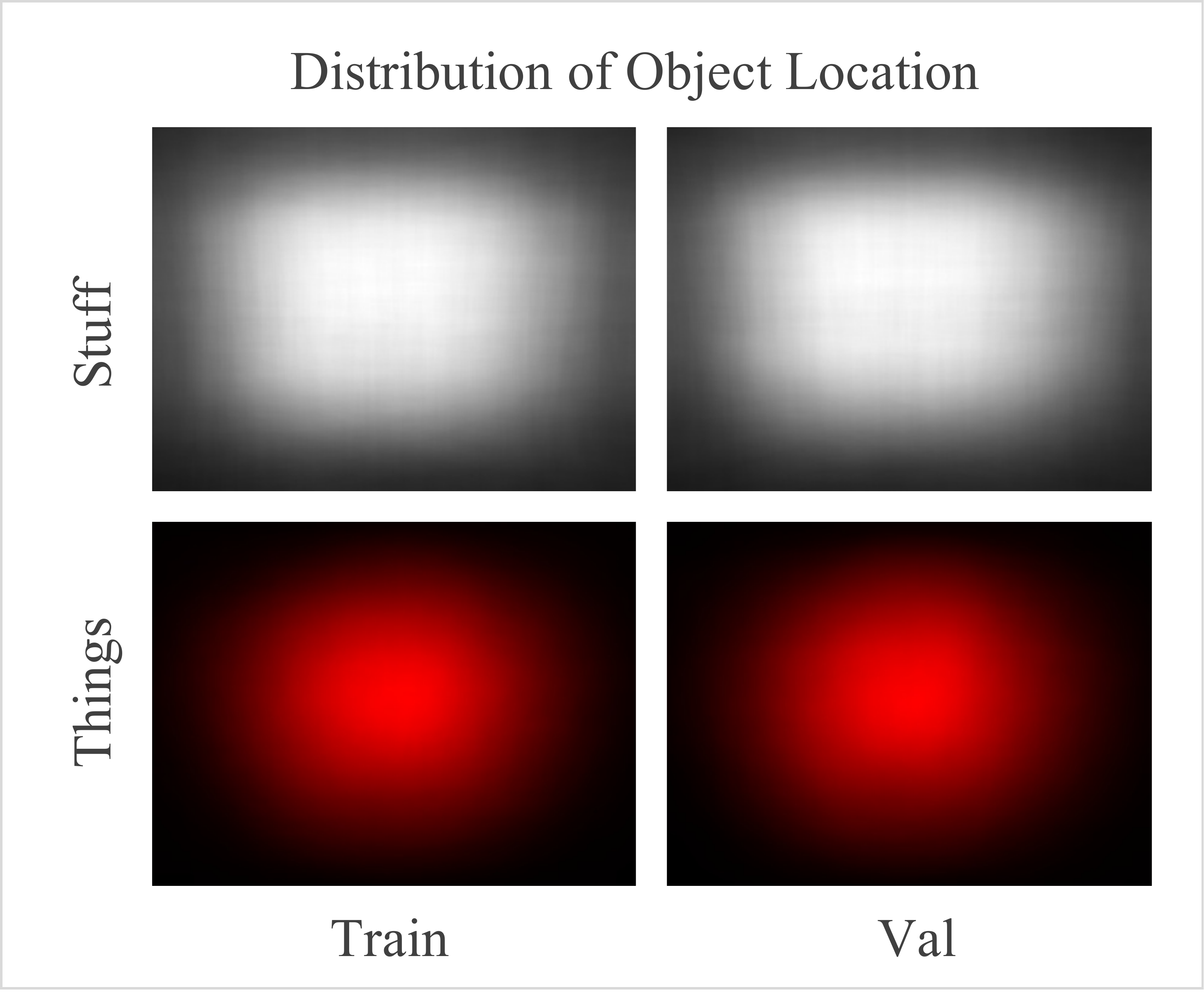}}
    \caption{
    Statistics of the Trans10K dataset.
    }
  \label{fig:statistic}
    \end{figure}

	\subsection{Evaluation Metrics}
	For a comprehensive evaluation, we apply four metrics that are widely used in semantic segmentation, salient object detection and shadow detection to benchmark the performance of transparent object segmentation. Our metrics are far more comprehensive than TransCut and TOM-Net, we hope they can expose more flaws of different methods on our dataset.
    Specifically, Intersection over union~\textbf{(IoU)} and pixel accuracy metrics~\textbf{(Acc)} are used from the semantic segmentation field as our first and second metrics. Note that we only calculate IoU of the thing and stuff, ignoring the background.
    Mean absolute error~\textbf{(MAE)} metrics are used from the salient object detection field.
    Finally, Balance error rate~\textbf{(BER)} is used from the shadow detection field. It considers the unbalanced areas of transparent and non-transparent regions.
    BER is used to evaluate binary predictions, here we change it to mean balance error rate~\textbf{(mBER)} to evaluate the two fine-grained transparent categories.
	, it is computed as:
	\begin{equation}
	mBER = \frac{1}C \sum\nolimits_{i = 1}^{C} (1-\frac{1}2(\frac{TP_{i}}{TP_{i}+FN_{i}} + \frac{TN_{i}}{TN_{i}+FP_{i}}))\times 100,
	\label{eqn:r_i}
	\end{equation}
	Where $C$ is the category of transparent objects, in this dataset $C$ is 2.

     \begin{figure}[ht]
    \centering
    \includegraphics[width=0.95\textwidth]{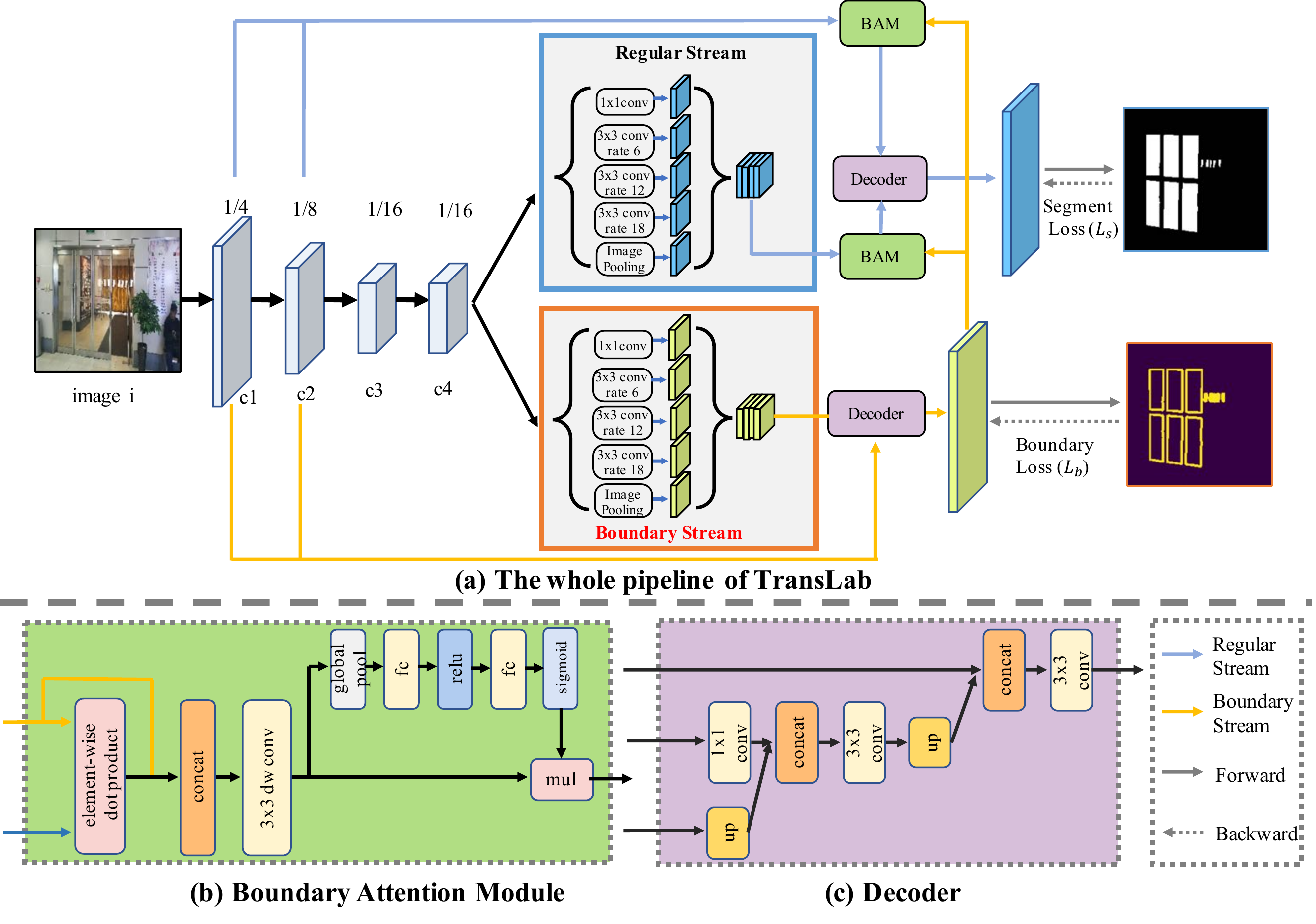}
    \caption{The architecture of TransLab.}
    \label{fig:pipeline}
\end{figure}

	\section{Proposed Method}
	\subsection{Network Architecture} Fig.~\ref{fig:pipeline}~(a) shows the overall architecture of TransLab, which is composed of two parallel stream: regular stream and boundary stream. ResNet50~\cite{resnet} with dilation is used as the backbone network. The regular stream is for transparent object segmentation while the boundary stream is for boundary prediction.
    We argue that the boundary is easier than content to observe because it tends to have high contrast in the edge of transparent objects, which is consistent with human visual perception.
    So we first make the network predict the boundary, then we utilize the predicted boundary map as a clue to attend the regular stream. It is implemented by Boundary Attention Module~(BAM).
    In each stream, we use Atrous Spatial Pyramid Pooling module~(ASPP) to enlarge the receive field.
    Finally, we design a simple decoder to utilize both high-level feature~(C4) and low-level feature~(C1 and C2).

	\subsection{Boundary Attention Module}
	Fig.~\ref{fig:pipeline}~(b) illustrates the structure of the Boundary Attention Module~(BAM).
    The boundary ground-truth is generated as a binary map with thickness 8. The channel of the predicted boundary map is 1.
    BAM first takes the feature map of the regular stream and the predicted boundary map as input. Then performs boundary attention with a boundary map. The two feature maps before and after boundary attention are concatenated and followed by a channel attention block. Finally, it outputs the refined feature map.
    BAM can be repeatedly used on the high-level and low-level features of regular stream such as C1, C2 and C4. We demonstrate that the more times of boundary attention, the better performance of segmentation results. Details are shown in Section~5.

    \subsection{Decoder}
    Fig.~\ref{fig:pipeline}~(c) illustrates the detailed structure of the Decoder. The input of decoder is C1, C2 and  C4~(after ASPP). BAM is used to apply boundary attention to three of them.
    We firstly fuse the C4 and C2 by up-sampling C4 and adding 3$\times$3 convolutional operation. The fused feature map is further up-sampled to fuse with C1 in the same approach.
    In this way, both high level and low-level feature maps are joint fused, which is beneficial for semantic segmentation.

	\subsection{Loss Function}
	We define our training loss function as follows:
	\begin{equation}
	L = L_s +  \lambda L_b,
	\label{eqn:loss-tot}
	\end{equation}
	where $L_s$ and $L_b$ represent the losses for the segmentation text and boundary, respectively, and $\lambda$ balances the importance between $L_s$ and $L_b$. The trade-off parameter $\lambda$ is set to 5 in our experiments.
	Here $L_s$ is the standard Cross-Entropy~(CE) Loss.
	Inspired by~\cite{milletari2016v}, we adopt Dice Loss in our experiment. The dice coefficient $D(S_i, G_i)$ is formulated as in Eqn.~\eqref{eqn:dice-coef}:
	\begin{equation}
	D(S_i, G_i) = \frac{2 \sum\nolimits_{x,y} (S_{i, x, y} \times G_{i, x, y})}{\sum\nolimits_{x, y} S_{i, x, y}^2 + \sum\nolimits_{x, y} G_{i, x, y}^2},
	\label{eqn:dice-coef}
	\end{equation}
	where $S_{i, x, y}$ and $G_{i, x, y}$ refer to the value of pixel $(x, y)$ in segmentation result $S_i$ and ground truth $G_i$, respectively.


    \section{Experiments}
    \subsection{Implementation Details}
    We have implemented TransLab with PyTorch~\cite{pytorch}. We use the pre-trained ResNet50~\cite{resnet} as the feature extraction network. In the final stage, we use the dilation convolution to keep the resolution as $1/16$. The remaining parts of our network are randomly initialized. For loss optimization, we use the stochastic gradient descent~(SGD) optimizer with momentum of 0.9 and a weight decay of 0.0005. Batch size is set to 8 per GPU. The learning rate is initialized to 0.02 and decayed by the poly strategy~\cite{bisenet} with the power of 0.9 for 16 epochs. We use 8 V100 GPU for all experiments.
    During training and inference, images are resized to a resolution of 512$\times$512.

    \subsection{Ablation Studies}
    In this part, we demonstrate the effectiveness of boundary clues for transparent object segmentation.
    We report the results of the ablation study on the \textbf{hard} set of Trans10K because it is more challenging and can clearly observe the gap of different modules.
    We simply use the DeeplabV3+ as our baseline.
    We first show only use boundary loss as auxiliary loss during training can directly improve the segmentation performance.
    We further show how to use a boundary map as a clue to attend the feature of regular stream. Experiments demonstrate that more boundary attention leading to better results. In summary, locating boundaries is essential for transparent object segmentation.

    \textbf{Boundary Loss Selection.}
    Boundary is easier to observe than the content of the transparent object.
    To obtain high-quality boundary map, the loss function is essential.
    We choose Binary Cross-Entropy Loss, Focal Loss and Dice Loss to supervise boundary stream.
    As shown in Table.~\ref{tab:ab2}, Firstly, simply using boundary loss as auxiliary loss can directly improve the performance no matter which loss function is used. We argue this is due to the benefit of multi-task learning. It means the boundary loss can help the backbone focus more on the boundary of transparent objects and extract more discriminative features. Note that under this setting, the boundary stream can be removed during inference so it will not bring computation overhead.
    Secondly, Focal Loss works better than Binary Cross-Entropy loss because the majority of pixels of a boundary mask are background, and the Focal Loss can mitigate the sample imbalance problem by decreasing the loss of easy samples. However, the Dice Loss achieves the best results without manually adjusting the loss hyper-parameters. Dice Loss views the pixels as a whole object and can establish the right balance between foreground and background pixels automatically. As a result, Dice Loss can improve baseline with 1.25\% on mIoU and 2.31\% on Acc, which is the best in three loss functions. Meanwhile, the mBer and MAE are also improved~(lower is better).

    \makeatletter
	\newcommand\figcaption{\def\@captype{figure}\caption}
	\newcommand\tabcaption{\def\@captype{table}\caption}
	\makeatother
	\begin{figure}[t]
		\begin{minipage}[t]{.45\linewidth}
			\centering
			\setlength{\tabcolsep}{1.2mm}
			\tabcaption{Ablation study for different loss functions of boundary stream.}
			\scalebox{0.8}{
			\begin{tabular}{p{30pt}<{\centering}|p{30pt}<{\centering}|p{30pt}<{\centering}|p{30pt}<{\centering}|p{30pt}<{\centering}}
	\hline
	\rule{0pt}{12pt}$L_{b}$ & mIoU~$\uparrow$ & Acc~$\uparrow$ & mBer~$\downarrow$ & MAE~$\downarrow$  \\
	\hline
	\rule{0pt}{12pt}- & 69.04 &78.07  &17.27 &0.194   \\
	\rule{0pt}{12pt}BCE & 69.33 &78.61 &16.89 &0.190   \\
	\rule{0pt}{12pt}Focal &69.41 &78.76 &16.27 &0.188  \\
	\hline
	\rule{0pt}{12pt}Dice &\textbf{70.29} &\textbf{80.38} &\textbf{15.44}  &\textbf{0.183}  \\
	\hline
\end{tabular}
			}

			\label{tab:ab2}
		\end{minipage}
		\begin{minipage}[t]{.50\linewidth}
			\centering
			\setlength{\tabcolsep}{1.2mm}
			\tabcaption{Ablation study for different setting of Boundary Attention Module.}
			\scalebox{0.8}{
			\begin{tabular}{p{60pt}<{\centering}|p{30pt}<{\centering}|p{30pt}<{\centering}|p{30pt}<{\centering}|p{30pt}<{\centering}}
	\hline
	\rule{0pt}{12pt}Method & mIoU~$\uparrow$ & Acc~$\uparrow$ & mBer~$\downarrow$ & MAE~$\downarrow$  \\
	\hline
	\rule{0pt}{12pt}BL &70.29 &80.38 &15.44 &0.183 \\
	\rule{0pt}{12pt}BL+C1 &71.05 &81.80 &13.98 &0.180   \\
	\rule{0pt}{12pt}BL+C1\&2 &71.32 &82.05 &13.69 &0.178   \\
	
	\hline
    \rule{0pt}{12pt}	BL+C1\&2\&4 &\textbf{72.10} &\textbf{83.04} &\textbf{13.30} &\textbf{0.166}  \\
	\hline
\end{tabular}

			}

			\label{tab:ab1}
		\end{minipage}
	\end{figure}

    \textbf{Boundary Attention Module.}
    After obtaining a high-quality boundary map, boundary attention is another key step in our algorithm.
    We can fuse boundary information at different levels for the regular stream.
    In this part, we repeatedly use BAM on C1, C2, C4 feature maps to show how boundary attention module works.
    To evaluate the relationship between the quantity of boundary attention and the final prediction accuracy, we vary the number of fusion levels from 1 to 3 and report the mIoU, Acc, mBer and MAE in Table~\ref{tab:ab1}.
    Note that `BL' indicates baseline with Dice Loss.
    It can be observed that performance is improved consistently by using boundary attention module at more levels.
    Our final model that fuses boundary information in all three levels further improves mIoU from 70.29\% to 72.10\% and Acc from 80.38\% to 83.04\%. Meanwhile, the mBer and MAE are also improved~(lower is better).
    By using a high-quality boundary map for attention, the feature maps of regular stream can have higher weights on the boundary region.

    \subsection{Comparison to the State-of-the-arts}
    We select several main-stream semantic segmentation methods to evaluate on our challenging Trans10K dataset.
    Specifically, we choose FPENet~\cite{fpenet}, ContextNet~\cite{contextnet}, FastSCNN~\cite{fastscnn}, DeeplabV3+ with MobilenetV2~\cite{deeplabv3+}, CGNet~\cite{cgnet}, HRNet~\cite{hrnet}, HardNet~\cite{hardnet}, DABNet~\cite{dabnet}, LEDNet~\cite{lednet}, ICNet~\cite{icnet} and BiSeNet~\cite{bisenet} as real-time methods.
    DenseASPP~\cite{denseaspp}, DeepLabV3+ with Resnet50~\cite{deeplabv3+},  FCN~\cite{fcn}, OCNet~\cite{ocnet}, RefineNet~\cite{refinenet}, DeepLabV3+ with Xception65~\cite{deeplabv3+}, DUNet~\cite{dunet}, UNet~\cite{unet} and PSPNet~\cite{pspnet} as regular methods.
    We re-train each of the models on the training set of our dataset and evaluate them on our testing set. For a fair comparison, we set the size of the input image as 512$\times$512 with single scale training and testing.

\begin{figure}[ht]
        \centering
        \scalebox{1}{
        \includegraphics[width=0.95\textwidth]{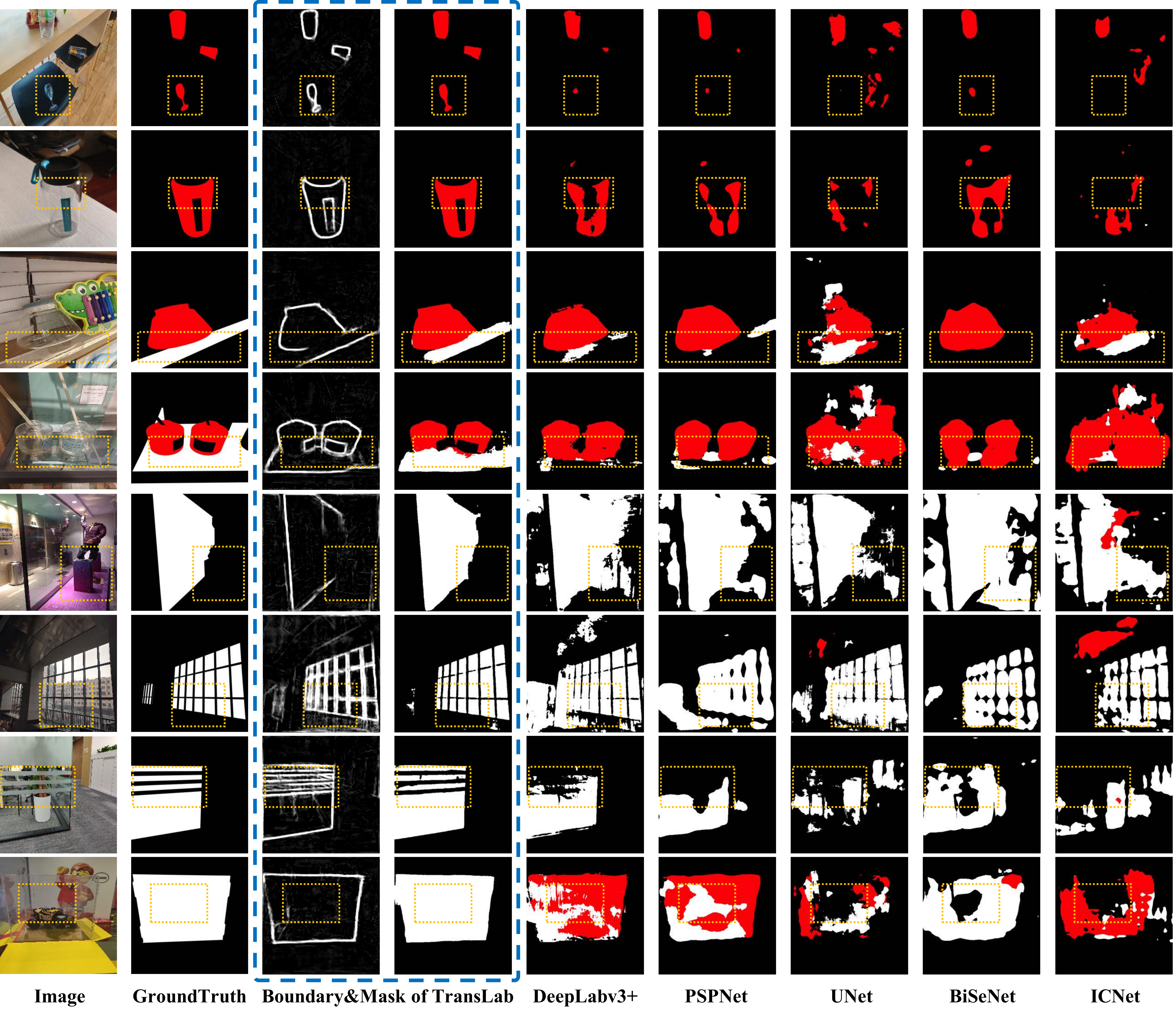}
        }
        \caption{Visual comparison of TransLab to other semantic segmentation methods. Our TransLab clearly outperforms others thanks to the boundary attention, especially in yellow dash region.}
        \label{fig:res}
    \end{figure}

    Table~\ref{tab:my_label} reports the overall quantitative comparison results on test set, where our TransLab outperforms all other methods in our Trans10K in terms of all four metrics in both easy/hard set and things/stuff categories.
    Especially, TransLab outperforms DeepLabV3+, the sota semantic segmentation method, in a large gap on all metrics as well as both things and stuff, especially in hard set.
    For instance, it surpasses DeepLabV3 by 3.97\% on `Acc'~(hard set). Fig~\ref{fig:res} also shows TransLab can predict sharp boundary with high-quality masks when compared with other methods. More analysis are shown in supplementary material.

    \begin{table}[ht]
    \centering
    \caption{Evaluated Semantic Segmentation methods. \textbf{Sorted by FLOPs}. Note that FLOPs is computed with one 512$\times$512 image.}

    \begin{center}
    (a) Comparison between things and stuff.
    \end{center}
    \begin{minipage}[ht]{0cm}
    \centerline{
    \scalebox{0.8}{
    \begin{tabular}{p{100pt}<{\centering}|p{37pt}<{\centering}|p{37pt}<{\centering}|p{37pt}<{\centering}p{37pt}<{\centering}|p{37pt}<{\centering}p{37pt}<{\centering}|p{37pt}<{\centering}p{37pt}<{\centering}}
\hline
\multirow{2}{*}{\large Method} & \multirow{2}{*}{MAE~$\downarrow$} & \multirow{2}{*}{ACC~$\uparrow$} & \multicolumn{2}{c|}{\rule{0pt}{10pt}IoU~$\uparrow$} & \multicolumn{2}{c|}{BER~$\downarrow$} & 
\multicolumn{2}{c}{Computation} \\ \cline{4-9} 
 & \rule{0pt}{10pt} & & Things & Stuff & Things & Stuff & FLOPs & Params \\ \hline
 \rule{0pt}{10pt} FPENet~\cite{fpenet} & 0.339 & 24.73 & 33.96 & 34.36 & 39.59 & 39.01 & 0.78G & 0.11M  \\
\rule{0pt}{10pt} ContextNet~\cite{contextnet} & 0.217 & 62.09 & 56.29 & 56.61 & 22.26 & 22.46 & 0.89G & 0.87M  \\
\rule{0pt}{10pt} FastSCNN~\cite{fastscnn} & 0.206 & 64.20 & 58.62 & 59.74 & 20.59 & 23.95 & 1.03G & 1.20M  \\
\rule{0pt}{10pt} DeepLabv3+MBv2~\cite{deeplabv3+} & 0.130 & 80.92 & 78.55 & 71.97 & 10.38 & 14.58 & 2.70G & 1.96M  \\
\rule{0pt}{10pt} CGNet~\cite{cgnet} & 0.216 & 59.15 & 58.33 & 56.28 & 21.02 & 24.88 & 3.52G & 0.49M  \\
\rule{0pt}{10pt} HRNet~\cite{hrnet} & 0.134 & 75.82 & 79.34 & 69.78 & 10.39 & 16.64 & 4.20G & 1.53M  \\
\rule{0pt}{10pt} HardNet~\cite{hardnet} & 0.184 & 69.17 & 64.91 & 63.15 & 17.18 & 20.63 & 4.43G & 4.11M  \\
\rule{0pt}{10pt} DABNet~\cite{dabnet} & 0.230 & 54.87 & 54.48 & 55.32 & 25.77 & 25.64 & 5.25G & 0.75M  \\
\rule{0pt}{10pt} LEDNet~\cite{lednet} & 0.168 & 75.70 & 70.37 & 64.68 & 12.68 & 17.62 & 6.32G & 2.31M  \\
\rule{0pt}{10pt} ICNet~\cite{icnet} & 0.244 & 52.65 & 53.90 & 47.38 & 19.78 & 29.46 & 10.66G & 8.46M  \\
\rule{0pt}{10pt} BiSeNet~\cite{bisenet} & 0.140 & 77.92 & 77.39 & 70.46 & 10.86 & 17.04 & 19.95G & 13.30M  \\

\hline
\rule{0pt}{10pt} DenseASPP~\cite{denseaspp} & 0.114 & 81.22 & 81.79 & 74.41 & 9.07 & 15.31 & 36.31G & 29.09M \\
\rule{0pt}{10pt} DeepLabv3+R50~\cite{deeplabv3+} & 0.081 & 89.54 & 87.90 & 81.16 & 5.31 & 10.25 & 37.98G & 28.74M  \\
\rule{0pt}{10pt} FCN~\cite{fcn} & 0.108 & 83.79 & 84.40 & 74.92 & 7.30 & 13.36 & 42.35G & 34.99M  \\
\rule{0pt}{10pt} OCNet~\cite{ocnet} & 0.122 & 80.85 & 80.55 & 73.15 & 8.91 & 16.38 & 43.43G & 35.91M  \\
\rule{0pt}{10pt} RefineNet~\cite{refinenet} & 0.180 & 57.97 & 73.65 & 58.40 & 16.44 & 27.98 & 44.34G & 29.36M  \\
\rule{0pt}{10pt} DeepLabv3+XP65~\cite{deeplabv3+} & 0.082 & 89.18 & 87.54 & 80.98 & 5.64 & 10.34 & 51.95G & 41.05M  \\
\rule{0pt}{10pt} DUNet~\cite{dunet} & 0.140 & 77.84 & 79.10 & 69.00 & 10.53 & 15.84 & 123.35G & 31.21M  \\
\rule{0pt}{10pt} UNet~\cite{unet} & 0.234 & 51.07 & 54.99 & 52.96 & 27.04 & 25.69 & 124.62G & 13.39M  \\
\rule{0pt}{10pt} PSPNet~\cite{pspnet} & 0.093 & 86.25 & 86.13 & 78.42 & 6.68 & 12.75 & 187.27G & 50.99M  \\

\hline \hline
\rule{0pt}{10pt} \textbf{TransLab} & \textbf{0.063} & \textbf{92.69} & \textbf{90.87} & \textbf{84.39} & \textbf{3.63} & \textbf{7.28} & 61.60G & 42.19M \\ 
\hline
\end{tabular}}
    }
    \end{minipage}
    \begin{center}(b) Comparison between Easy and Hard.\end{center}
    \begin{minipage}[ht]{8cm}
    \centerline{
    \scalebox{0.7}{
    \begin{tabular}{p{95pt}<{\centering}|p{31pt}<{\centering}p{31pt}<{\centering}p{31pt}<{\centering}|p{31pt}<{\centering}p{31pt}<{\centering}p{31pt}<{\centering}|p{31pt}<{\centering}p{31pt}<{\centering}p{31pt}<{\centering}|p{31pt}<{\centering}p{31pt}<{\centering}p{31pt}<{\centering}}
\hline\multirow{2}{*}{\large Method} & \multicolumn{3}{c|}{\rule{0pt}{10pt}MAE~$\downarrow$} & \multicolumn{3}{c|}{Acc~$\uparrow$} & \multicolumn{3}{c|}{mIoU~$\uparrow$} & \multicolumn{3}{c}{mBER~$\downarrow$} \\ \cline{2-13} 
 & \rule{0pt}{10pt}All & Easy & Hard & All & Easy & Hard & All & Easy & Hard & All & Easy & Hard \\ \hline
\rule{0pt}{10pt} FPENet~\cite{fpenet} & 0.339 & 0.297 & 0.492 & 24.73 & 26.50 & 19.19 & 34.17 & 36.82 & 24.41 & 39.31 & 37.88 & 44.03 \\
\rule{0pt}{10pt} ContextNet~\cite{contextnet} & 0.217 & 0.171 & 0.386 & 62.09 & 67.14 & 46.34 & 56.46 & 61.73 & 37.71 & 22.36 & 18.77 & 34.44 \\
\rule{0pt}{10pt} FastSCNN~\cite{fastscnn} & 0.206 & 0.161 & 0.373 & 64.20 & 69.42 & 48.01 & 59.18 & 64.63 & 40.27 & 22.27 & 18.74 & 34.22 \\
\rule{0pt}{10pt} DeepLabv3+MBv2\cite{deeplabv3+} & 0.130 & 0.091 & 0.275 & 80.92 & 85.90 & 65.43 & 75.27 & 80.55 & 56.17 & 12.49 & 9.08 & 24.47 \\
\rule{0pt}{10pt} CGNet~\cite{cgnet} & 0.216 & 0.173 & 0.379 & 59.15 & 64.57 & 42.26 & 57.31 & 62.41 & 39.56 & 22.95 & 19.67 & 34.33 \\ 
\rule{0pt}{10pt} HRNet~\cite{hrnet} & 0.134 & 0.092 & 0.291 & 75.82 & 82.17 & 56.04 & 74.56 & 80.43 & 53.42 & 13.52 & 9.95 & 26.17 \\
\rule{0pt}{10pt} HardNet~\cite{hardnet} & 0.184 & 0.141 & 0.345 & 69.17 & 73.83 & 54.67 & 64.03 & 69.11 & 46.18 & 18.91 & 15.58 & 30.52 \\
\rule{0pt}{10pt} DABNet~\cite{dabnet} & 0.230 & 0.187 & 0.391 & 54.87 & 59.29 & 41.07 & 54.90 & 59.45 & 38.77 & 25.71 & 22.63 & 36.15 \\
\rule{0pt}{10pt}LEDNet~\cite{lednet} & 0.168 & 0.124 & 0.331 & 75.70 & 80.62 & 60.37 & 67.54 & 73.04 & 48.38 & 15.15 & 11.83 & 26.58 \\
\rule{0pt}{10pt} ICNet~\cite{icnet} & 0.244 & 0.200 & 0.408 & 52.65 & 58.31 & 35.01 & 50.65 & 55.48 & 33.44 & 24.63 & 21.71 & 35.24 \\
\rule{0pt}{10pt} BiSeNet~\cite{bisenet} & 0.140 & 0.102 & 0.282 & 77.92 & 82.79 & 62.72 & 73.93 & 78.74 & 56.37 & 13.96 & 10.83 & 24.85 \\
\hline

\rule{0pt}{10pt} DenseAspp~\cite{denseaspp} & 0.114 & 0.078 & 0.247 & 81.22 & 86.25 & 66.55 & 78.11 & 83.11 & 60.38 & 12.19 & 8.85 & 23.71 \\
\rule{0pt}{10pt} DeepLabv3+R50\cite{deeplabv3+} & 0.081 & 0.050 & 0.194 & 89.54 & 93.22 & 78.07 & 84.54 & 89.09 & 69.04 & 7.78 & 4.91 & 17.27 \\
\rule{0pt}{12pt} FCN~\cite{fcn} &  0.108 & 0.073  & 0.239 & 83.79  &  88.55 & 68.93  & 79.67 & 84.53  & 62.51 & 10.33 & 7.36  & 20.47    \\
\rule{0pt}{12pt} OCNet~\cite{ocnet} & 0.122  & 0.087  & 0.253 & 80.85 & 85.63  & 65.96  &  76.85 & 81.53 & 59.75  & 12.65  &  9.43 & 23.69   \\ 
\rule{0pt}{12pt} RefineNet~\cite{refinenet} & 0.180  & 0.135  & 0.345 & 57.97 & 64.53 & 37.53  & 66.03 & 71.41  & 45.71 & 22.22  &  19.01 & 34.06   \\ 
\rule{0pt}{10pt} DeepLabv3+XP65~\cite{deeplabv3+} & 0.082 & 0.051 & 0.195 & 89.18 & 92.61 & 78.51 & 84.26 & 88.87 & 68.34 & 8.00 & 5.16 & 17.44 \\
\rule{0pt}{10pt} DUNet~\cite{dunet} & 0.140 & 0.100 & 0.289 & 77.84 & 83.41 & 60.50 & 74.06 & 79.19 & 55.53 & 13.19 & 9.93 & 25.01 \\
\rule{0pt}{12pt} UNet~\cite{unet} & 0.234 & 0.191  & 0.398  & 51.07  & 55.44 & 37.44  & 53.98  & 58.60 & 37.08  & 26.37  & 23.40 & 36.80    \\
\rule{0pt}{12pt} PSPNet~\cite{pspnet} & 0.093 & 0.062  & 0.211  & 86.25  & 90.41  & 73.28  & 82.38  & 86.79  & 66.35  & 9.72  & 6.67 &  20.08   \\

\hline \hline
\rule{0pt}{10pt} \textbf{TransLab} & \textbf{0.063} & \textbf{0.036} & \textbf{0.166} & \textbf{92.69} & \textbf{95.77} & \textbf{83.04} & \textbf{87.63} & \textbf{92.23} & \textbf{72.10} & \textbf{5.46} & \textbf{3.12} & \textbf{13.30}
 \\ \hline
\end{tabular}
    }}

    \end{minipage}
    \label{tab:my_label}
    \end{table}

    \section{Conclusion}
    In this work, we present the Trans10K dataset, which is the largest real dataset for transparent object segmentation.
    We also benchmark 20 semantic segmentation algorithms on this novel dataset and shed light on what attributes are especially difficult for current methods. We suggest that transparent object segmentation in the wild is far from being solved. Finally, we propose a boundary-aware algorithm, termed TransLab, to utilize the boundary prediction to improve the segmentation performance. \\
    \textbf{Acknowledgement}
This work is partially supported by the SenseTime Donation for Research, HKU Seed Fund for Basic Research, Startup Fund and General Research Fund No.27208720. Chunhua Shen and his employer received no financial support for the research, authorship, and/or publication of this article.

    \clearpage

    \appendix
    \section{Appendix}
    \subsection{Detailed annotation information.}
\begin{figure}
        \centering
        \subfigure[Histogram of image numbers for different transparent object categories. It is split by things and stuff. Event categories are ranked in an descending order based on the image numbers. Example images for specific transparent object classes are shown. Y-axis denotes for image numbers. X-axis denotes for transparent object categories.]{\includegraphics[width=1\textwidth]{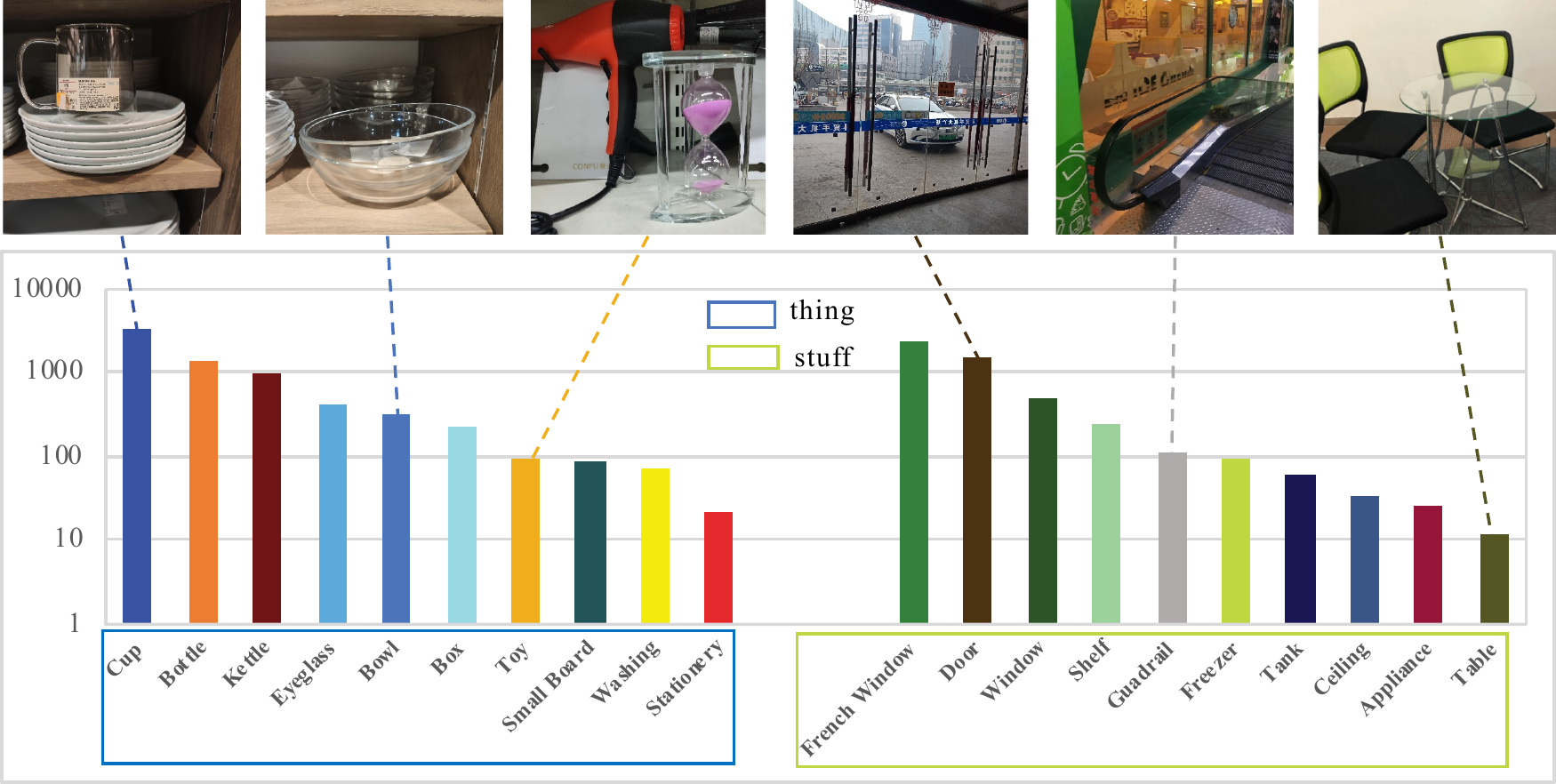}}
        \subfigure[Histogram of image numbers for different scene categories. Event categories are ranked in an descending order based on the image numbers. Example images for specific scene classes are shown. Y-axis denotes for image numbers. X-axis denotes for event scene name.]{\includegraphics[width=1\textwidth]{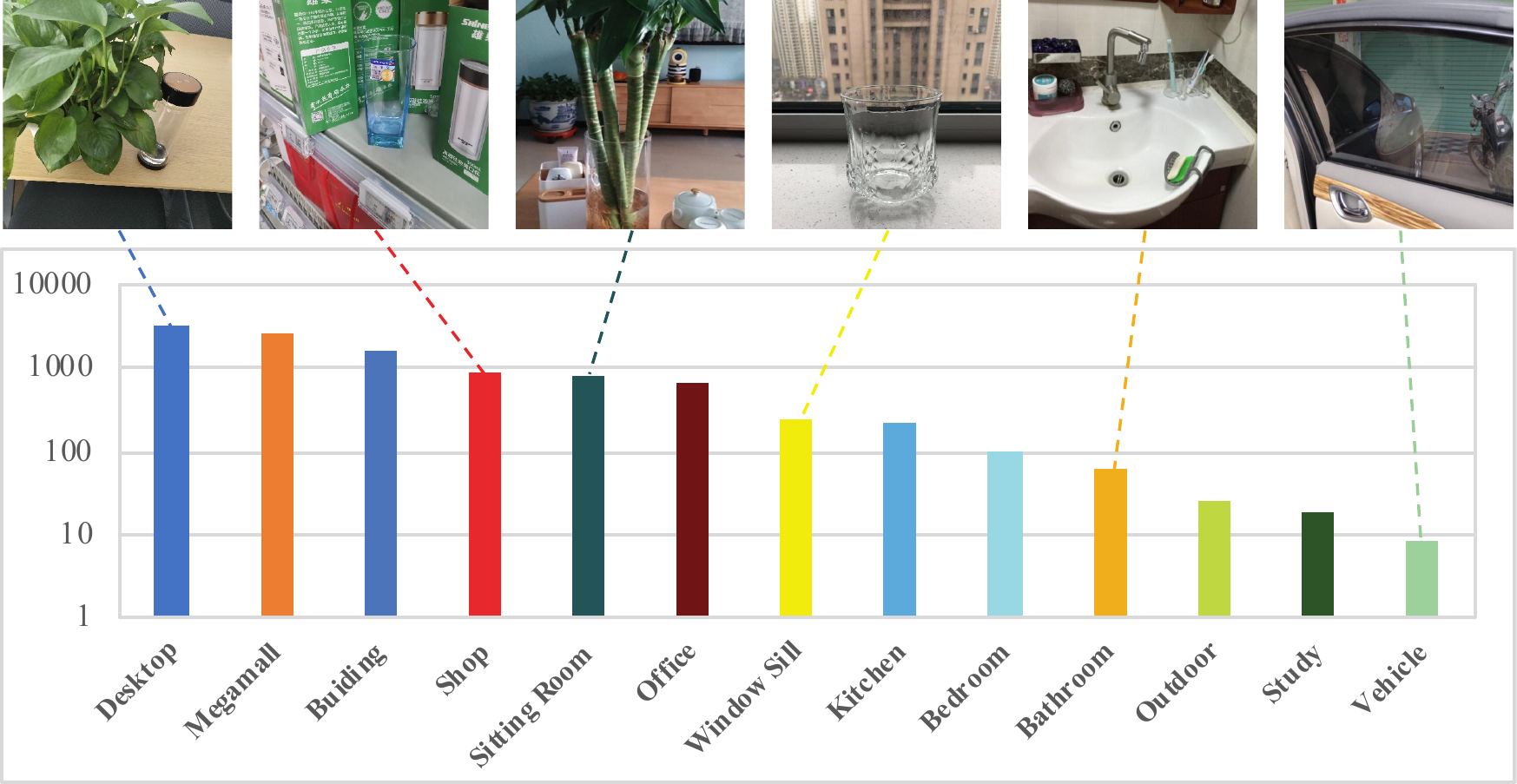}}
        \caption{Statistics of Trans10K dataset.}
        \label{fig:my_label}
    \end{figure}

    Our dataset contains 20 different categories of transparent objects. As shown in Fig.~\ref{fig:my_label}(a), we count the number of images for different categories. We see that for things, the ``Cup'' appears the most frequently  while the ``Stationery'' appears the least.
    For stuff, the ``French Window'' appears the most frequently while the ``Table'' appears the least.
    Furthermore, our dataset also contains 13 scenarios. As shown in Fig.~\ref{fig:my_label}(b), we see that the ``Desktop'' is the most frequently scene while the ``Vechile'' is the least scene.
    In summary, our Trans10K contains abundant category distribution of transparent objects and scenarios, which is not available in TOM-Net~\cite{tomnet} and TransCut~\cite{transcut}.

    \subsection{More visual results Analysis.}

    \subsubsection{Failure Samples Analysis}
\begin{figure}[ht]
    \centering
        \subfigure[High Transparency.]{\includegraphics[width=0.45\textwidth]{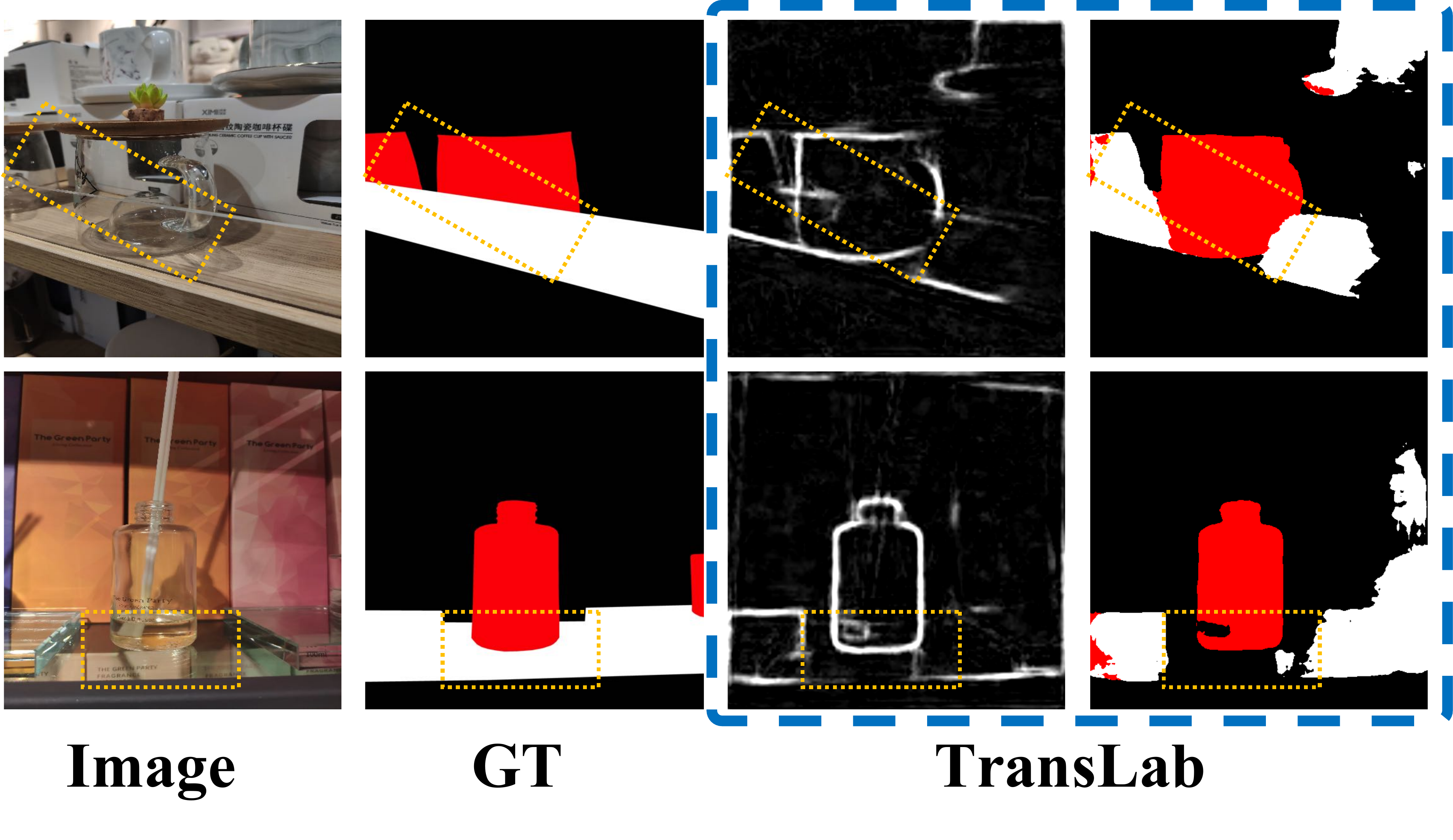}}
        \subfigure[Strong Reflection.]{\includegraphics[width=0.45\textwidth]{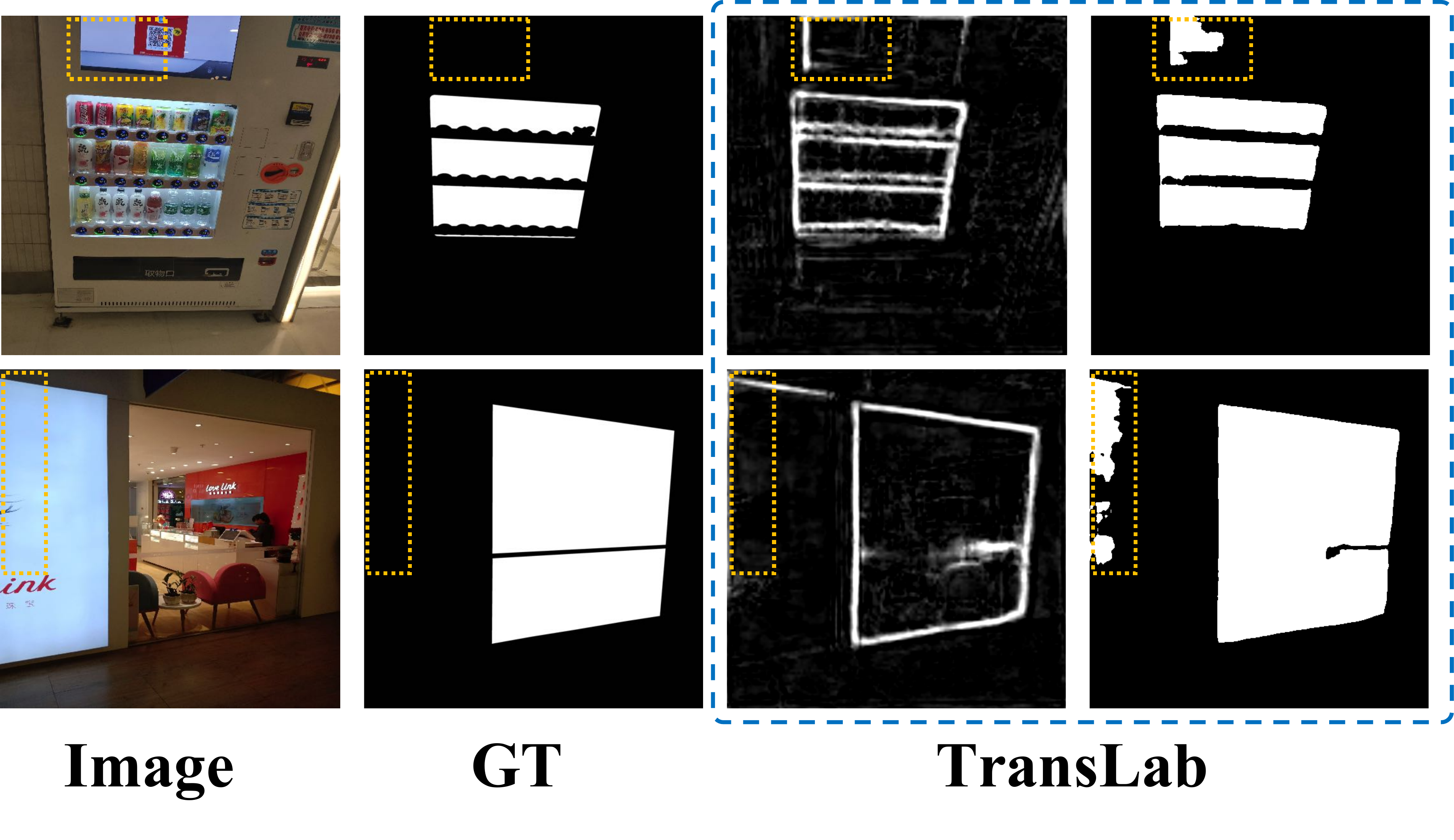}}
        \subfigure[Overlap \& Occlusion.]{\includegraphics[width=0.45\textwidth]{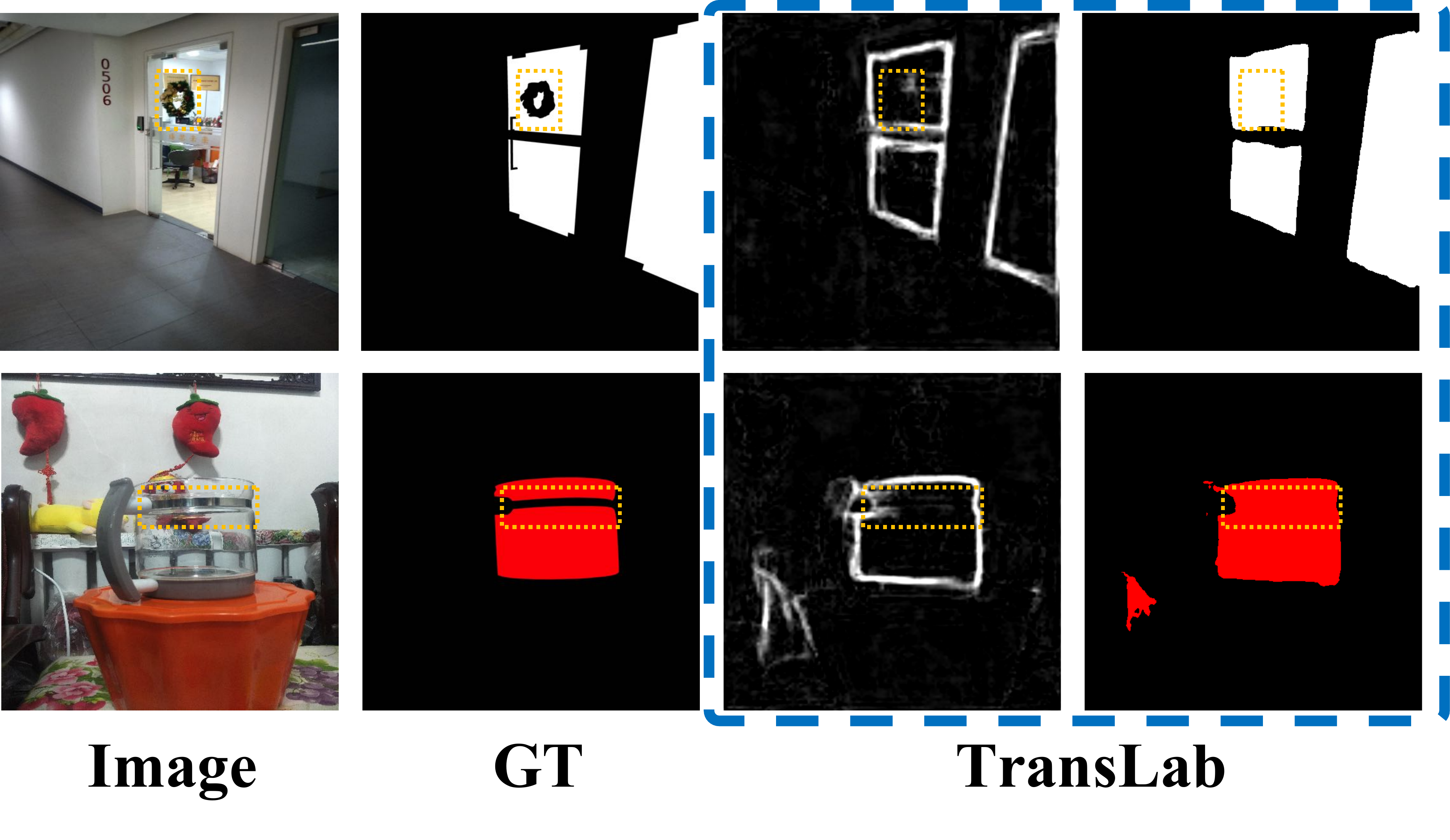}}
         \subfigure[Semi-Transparency.]{\includegraphics[width=0.45\textwidth]{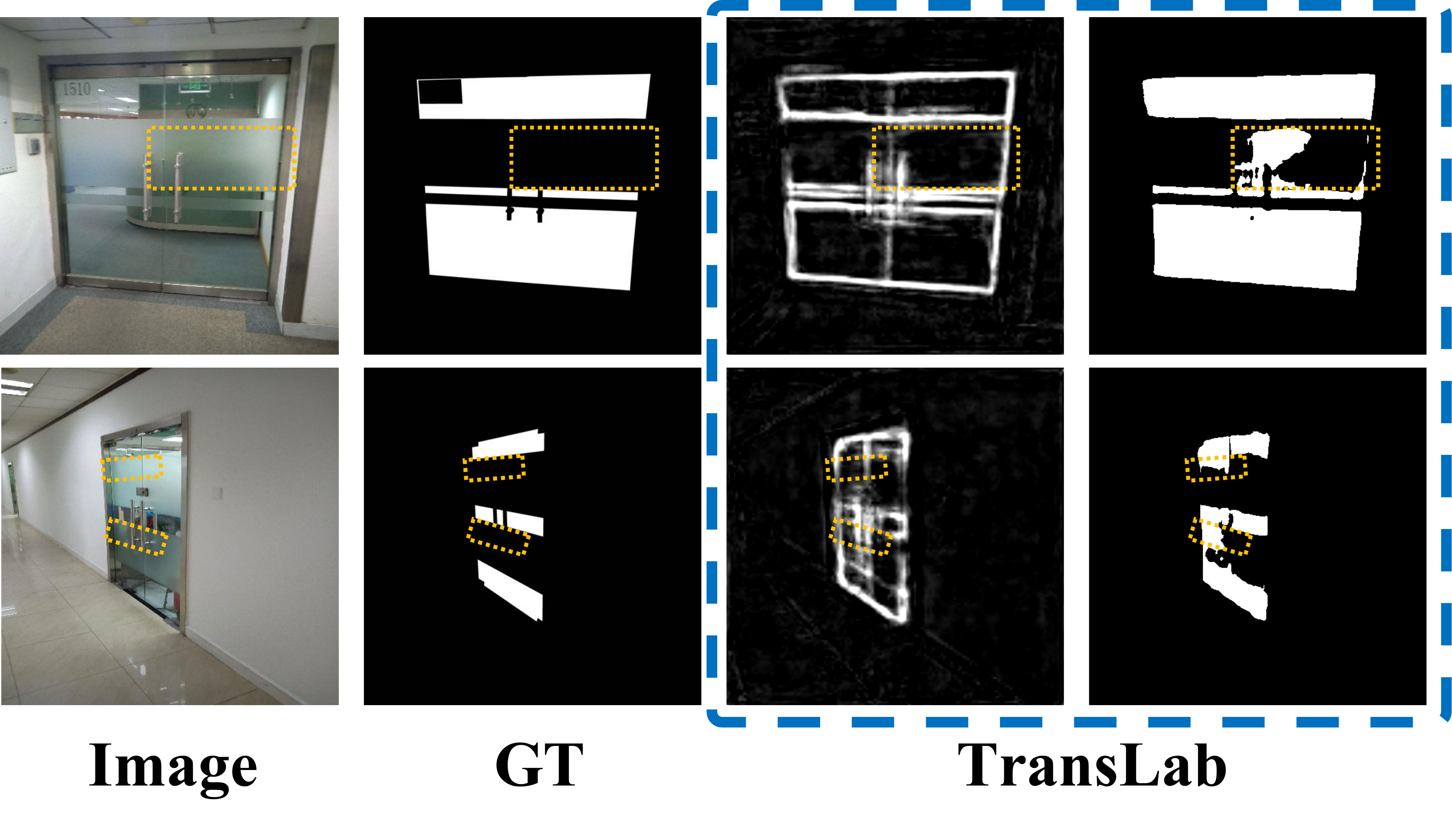}}
    \caption{Failure cases. Our method fails to segment transparent objects in some complex scenarios. }
    \label{fig:failure}
    \end{figure}

    As shown in Fig.~\ref{fig:failure}, our method has some limitations. For example, in Fig.~\ref{fig:failure}~(a), when facing highly-transparency objects, our method will fail to segment in some region. Fig.~\ref{fig:failure}~(b) shows that some objects with strong reflection will also make our method confuse and lead to wrong classification. In Fig.~\ref{fig:failure}~(c), we can find our method does not work well when some objects have overlap and occlusion with transparent objects. Finally, in Fig.~\ref{fig:failure}~(d), when semi-transparent objects are adjacent with transparent objects, our method will also confuse.

    \subsubsection{Visual results on external data.}
\begin{figure}[t]
    \centering
    \includegraphics[width=1\textwidth]{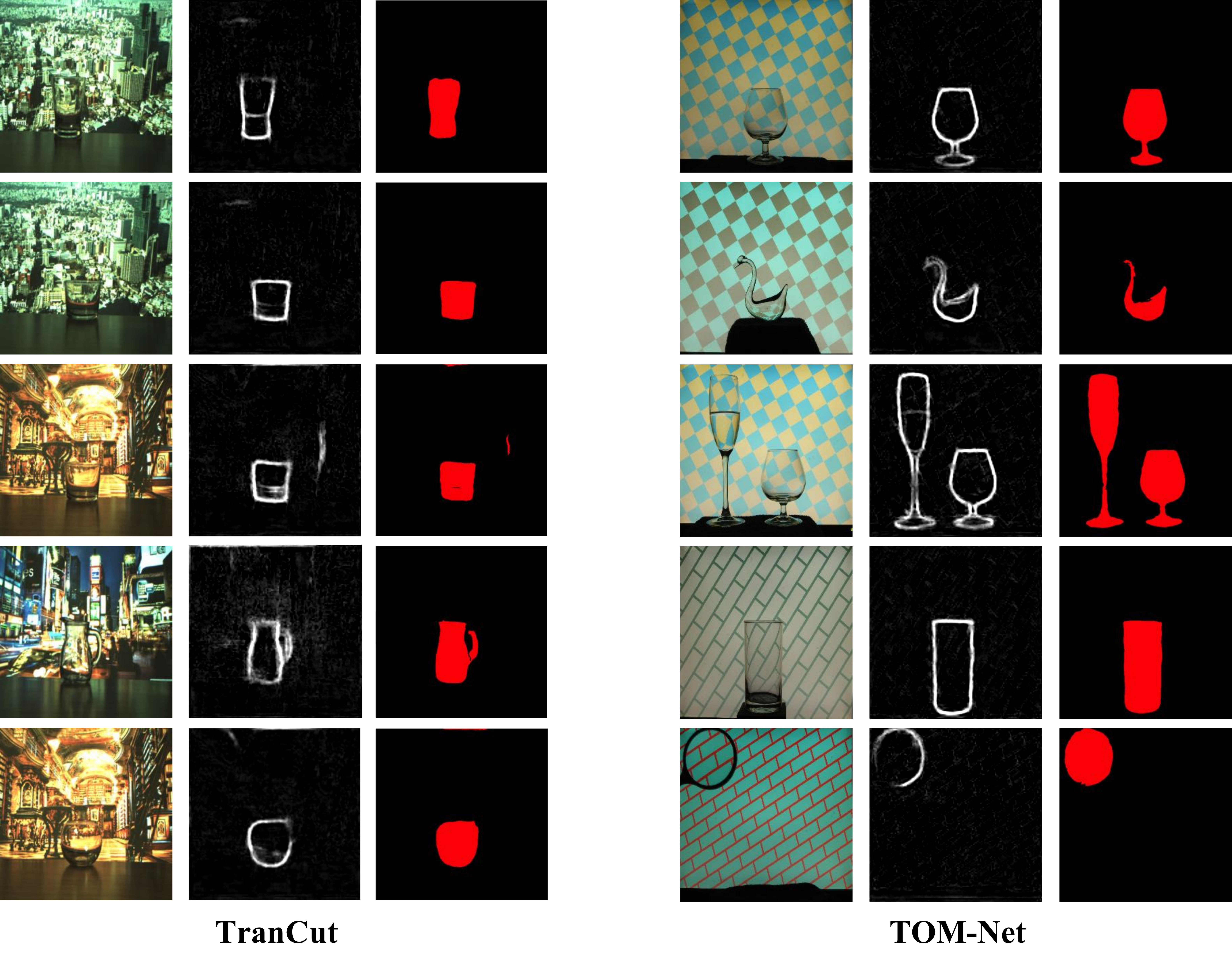}
    \vspace{-0.7cm}
    \caption{Some transparent objects segmentation results on two prior datasets: TransCut~\cite{transcut} and TOM-Net~\cite{tomnet}.}
    \label{fig:otherdata}
    \end{figure}

    In this part, we also directly test our TransLab trained on Trans10K dataset to evaluate the generalize ability of Trans10K dataset and the robustness of TransLab.
    Firstly, we test our method on two prior datasets: TransCut~\cite{transcut} and TOM-Net~\cite{tomnet}. As shown in Fig.~\ref{fig:otherdata}, our method can clearly output very high-quality mask.
    Moreover, we also test our method on some external data randomly captured by our mobile phones or obtained from Internet videos such as YouTube, TikTok and eBay.
    We see our TransLab can also successfully segment transparent objects in most cases.
    In summary, we believe our Trans10K dataset contains high-diversity images which can easily generalize to real scene. Also, our boundary-aware algorithm TransLab is robust enough to segment unseen images.

    \subsubsection{More comparison of TransLab with other methods.}
\begin{figure}[t]
    \centering
    \includegraphics[width=1\textwidth]{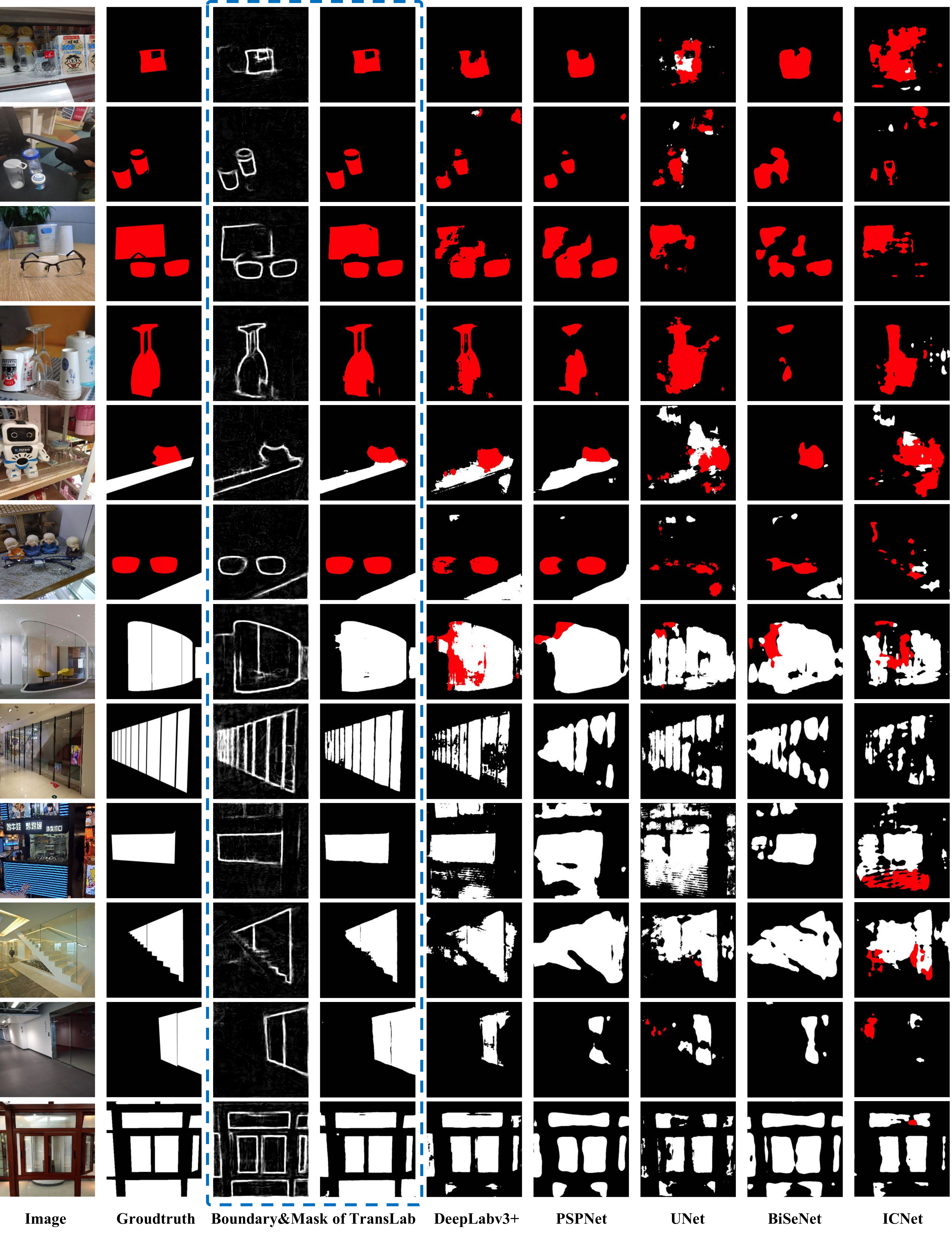}
    \vspace{-0.7cm}
    \caption{More visual comparison of TransLab to other semantic segmentation methods. Our TransLab clearly outperforms others thanks to the boundary attention.}
    \label{fig:supple1}
    \end{figure}

    In this part, we demonstrate more test examples produced by our TransLab on Trans10K dataset in Fig.~\ref{fig:supple1}. From these results, it can be easily observed that with the proposed Boundary Stream and Boundary Attention Module, our method can output high-quality boundary map and better transparent object segmentation mask than other semantic segmentation methods.

     \begin{figure}[t]
    \centering
    \includegraphics[width=1\textwidth]{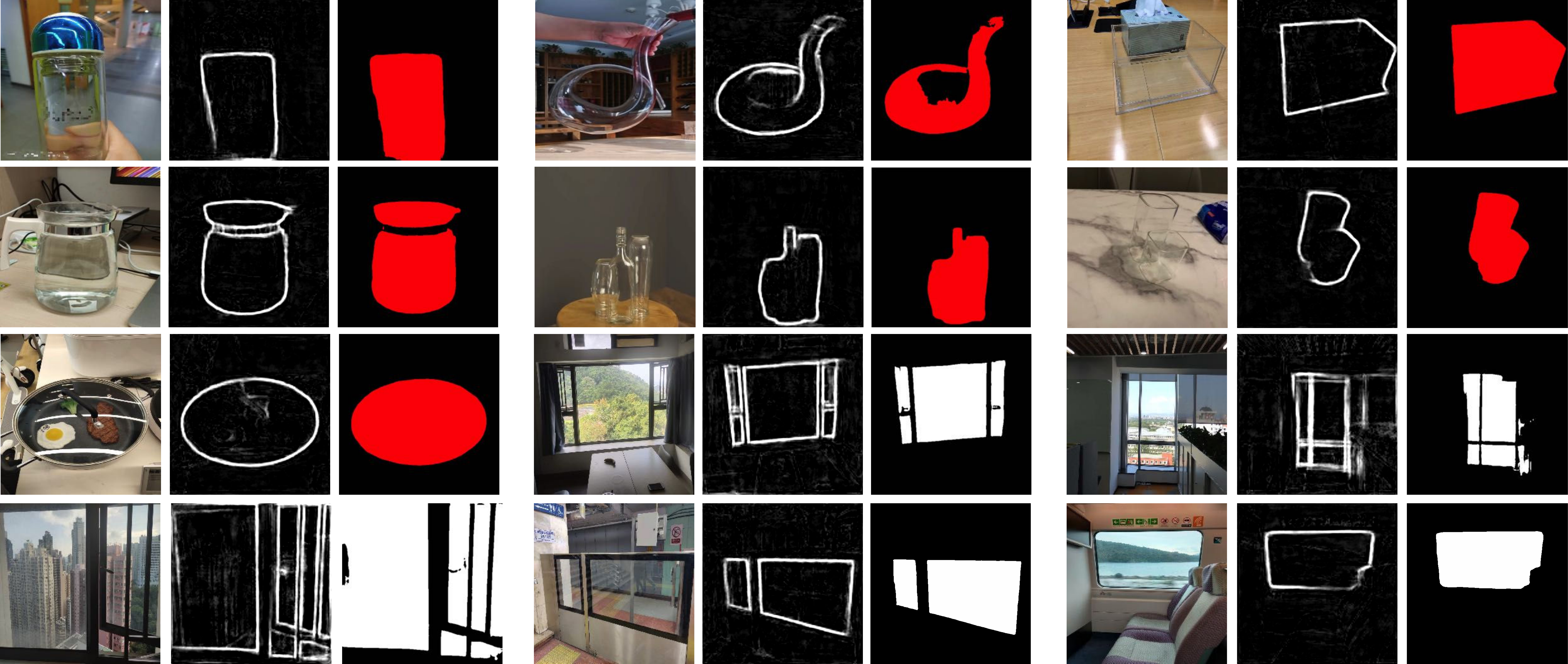}
    \vspace{-0.7cm}
    \caption{Some transparent objects segmentation results on challenging images captured by our mobile phones and obtained from Internet such as YouTube, TikTok and eBay.}
    \label{fig:internet}
    \end{figure}

    \clearpage

	\bibliographystyle{splncs}
	\bibliography{egbib}

\begin{thebibliography}{10}

\bibitem{transcut}
Xu, Y., Nagahara, H., Shimada, A., Taniguchi, R.:
\newblock Transcut: Transparent object segmentation from a light-field image.
\newblock In: ICCV. (2015)

\bibitem{tomnet}
Chen, G., Han, K., Wong, K.K.:
\newblock Tom-net: Learning transparent object matting from a single image.
\newblock In: CVPR. (2018)

\bibitem{pspnet}
Zhao, H., Shi, J., Qi, X., Wang, X., Jia, J.:
\newblock Pyramid scene parsing network.
\newblock In: CVPR. (2017)

\bibitem{icnet}
Zhao, H., Qi, X., Shen, X., Shi, J., Jia, J.:
\newblock Icnet for real-time semantic segmentation on high-resolution images.
\newblock In: ECCV. (2018)

\bibitem{dunet}
Jin, Q., Meng, Z., Pham, T.D., Chen, Q., Wei, L., Su, R.:
\newblock Dunet: A deformable network for retinal vessel segmentation.
\newblock Knowledge-Based Systems (2019)

\bibitem{deeplab}
Chen, L.C., Papandreou, G., Kokkinos, I., Murphy, K., Yuille, A.L.:
\newblock Deeplab: Semantic image segmentation with deep convolutional nets,
  atrous convolution, and fully connected crfs.
\newblock TPAMI (2017)

\bibitem{denseaspp}
Yang, M., Yu, K., Zhang, C., Li, Z., Yang, K.:
\newblock Denseaspp for semantic segmentation in street scenes.
\newblock In: CVPR. (2018)

\bibitem{refinenet}
Lin, G., Milan, A., Shen, C., Reid, I.:
\newblock Refinenet: Multi-path refinement networks for high-resolution
  semantic segmentation.
\newblock In: CVPR. (2017)

\bibitem{fcn}
Long, J., Shelhamer, E., Darrell, T.:
\newblock Fully convolutional networks for semantic segmentation.
\newblock In: CVPR. (2015)

\bibitem{deeplabv3+}
Chen, L.C., Zhu, Y., Papandreou, G., Schroff, F., Adam, H.:
\newblock Encoder-decoder with atrous separable convolution for semantic image
  segmentation.
\newblock In: ECCV. (2018)

\bibitem{crf}
Chen, L.C., Papandreou, G., Kokkinos, I., Murphy, K., Yuille, A.L.:
\newblock Semantic image segmentation with deep convolutional nets and fully
  connected crfs.
\newblock arXiv (2014)

\bibitem{lin2016efficient}
Lin, G., Shen, C., Van Den~Hengel, A., Reid, I.:
\newblock Efficient piecewise training of deep structured models for semantic
  segmentation.
\newblock In: CVPR. (2016)

\bibitem{zheng2015conditional}
Zheng, S., Jayasumana, S., Romera-Paredes, B., Vineet, V., Su, Z., Du, D.,
  Huang, C., Torr, P.H.:
\newblock Conditional random fields as recurrent neural networks.
\newblock In: ICCV. (2015)

\bibitem{chen2016semantic}
Chen, L.C., Barron, J.T., Papandreou, G., Murphy, K., Yuille, A.L.:
\newblock Semantic image segmentation with task-specific edge detection using
  cnns and a discriminatively trained domain transform.
\newblock In: CVPR. (2016)

\bibitem{deeplab2}
Chen, L.C., Zhu, Y., Papandreou, G., Schroff, F., Adam, H.:
\newblock Encoder-decoder with atrous separable convolution for semantic image
  segmentation.
\newblock In: ECCV. (2018)

\bibitem{gadde2016superpixel}
Gadde, R., Jampani, V., Kiefel, M., Kappler, D., Gehler, P.V.:
\newblock Superpixel convolutional networks using bilateral inceptions.
\newblock In: ECCV. (2016)

\bibitem{liu2017learning}
Liu, S., De~Mello, S., Gu, J., Zhong, G., Yang, M.H., Kautz, J.:
\newblock Learning affinity via spatial propagation networks.
\newblock In: NIPS. (2017)

\bibitem{nonlocal}
Wang, X., Girshick, R., Gupta, A., He, K.:
\newblock Non-local neural networks.
\newblock In: CVPR. (2018)

\bibitem{openimage}
Kuznetsova, A., Rom, H., Alldrin, N., Uijlings, J., Krasin, I., Pont-Tuset, J.,
  Kamali, S., Popov, S., Malloci, M., Duerig, T.,  et~al.:
\newblock The open images dataset v4: Unified image classification, object
  detection, and visual relationship detection at scale.
\newblock arXiv (2018)

\bibitem{resnet}
He, K., Zhang, X., Ren, S., Sun, J.:
\newblock Deep residual learning for image recognition.
\newblock In: CVPR. (2016)

\bibitem{milletari2016v}
Milletari, F., Navab, N., Ahmadi, S.A.:
\newblock V-net: Fully convolutional neural networks for volumetric medical
  image segmentation.
\newblock In: IC3DV. (2016)

\bibitem{pytorch}
Paszke, A., Gross, S., Chintala, S., Chanan, G., Yang, E., DeVito, Z., Lin, Z.,
  Desmaison, A., Antiga, L., Lerer, A.:
\newblock Automatic differentiation in pytorch.
\newblock (2017)

\bibitem{bisenet}
Yu, C., Wang, J., Peng, C., Gao, C., Yu, G., Sang, N.:
\newblock Bisenet: Bilateral segmentation network for real-time semantic
  segmentation.
\newblock In: ECCV. (2018)

\bibitem{fpenet}
Liu, M., Yin, H.:
\newblock Feature pyramid encoding network for real-time semantic segmentation.
\newblock arXiv (2019)

\bibitem{contextnet}
Poudel, R.P., Bonde, U., Liwicki, S., Zach, C.:
\newblock Contextnet: Exploring context and detail for semantic segmentation in
  real-time.
\newblock arXiv (2018)

\bibitem{fastscnn}
Poudel, R.P., Liwicki, S., Cipolla, R.:
\newblock Fast-scnn: fast semantic segmentation network.
\newblock arXiv (2019)

\bibitem{cgnet}
Wu, T., Tang, S., Zhang, R., Zhang, Y.:
\newblock Cgnet: A light-weight context guided network for semantic
  segmentation.
\newblock arXiv (2018)

\bibitem{hrnet}
Wang, J., Sun, K., Cheng, T., Jiang, B., Deng, C., Zhao, Y., Liu, D., Mu, Y.,
  Tan, M., Wang, X.,  et~al.:
\newblock Deep high-resolution representation learning for visual recognition.
\newblock arXiv (2019)

\bibitem{hardnet}
Chao, P., Kao, C.Y., Ruan, Y.S., Huang, C.H., Lin, Y.L.:
\newblock Hardnet: A low memory traffic network.
\newblock In: ICCV. (2019)

\bibitem{dabnet}
Li, G., Yun, I., Kim, J., Kim, J.:
\newblock Dabnet: Depth-wise asymmetric bottleneck for real-time semantic
  segmentation.
\newblock arXiv (2019)

\bibitem{lednet}
Wang, Y., Zhou, Q., Liu, J., Xiong, J., Gao, G., Wu, X., Latecki, L.J.:
\newblock Lednet: A lightweight encoder-decoder network for real-time semantic
  segmentation.
\newblock In: ICIP. (2019)

\bibitem{ocnet}
Yuan, Y., Wang, J.:
\newblock Ocnet: Object context network for scene parsing.
\newblock arXiv (2018)

\bibitem{unet}
Ronneberger, O., Fischer, P., Brox, T.:
\newblock U-net: Convolutional networks for biomedical image segmentation.
\newblock In: MICCAI. (2015)

\end{thebibliography}
\end{document}